\documentclass[10pt,twocolumn,letterpaper]{article}

\usepackage{xcolor}
\usepackage{cvpr} % uncomment this for the final submission
\definecolor{iccvblue}{rgb}{0.21,0.49,0.74}
\definecolor{iccvred}{rgb}{0.8,0.0,0.0} % Define red for figures and tables
\usepackage[pagebackref,breaklinks,colorlinks]{hyperref}
\hypersetup{
    linkcolor=blue,  % References (e.g., citations)
    citecolor=blue,  % Citation references
    filecolor=iccvblue,  % Files
    urlcolor=iccvblue,   % URLs
    allcolors=blue   % Default color
}

\AtBeginDocument{
    \hypersetup{
        linkcolor=red % Change figure and table references to red
    }
}

\usepackage{pifont}
\usepackage{booktabs}
\usepackage{array}
\usepackage{graphicx}
\usepackage{graphicx}
\usepackage{amsmath}
\usepackage{amssymb}
\usepackage{booktabs}
\usepackage{amsfonts,amssymb}
\usepackage{algorithm}
\usepackage{algpseudocode}
\usepackage{multirow}
\usepackage{float}
\usepackage{caption}
\usepackage{subcaption}
\usepackage{fontawesome}
\usepackage{tabularx,caption}
\usepackage[rightcaption]{sidecap}
\usepackage{colortbl}
\usepackage{psfrag}
\usepackage[percent]{overpic}
\usepackage{url}
\usepackage[accsupp]{axessibility}
\usepackage{tablefootnote}
\usepackage{caption} %
\newcommand{\tabincell}[2]{\begin{tabular}{@{}#1@{}}#2\end{tabular}}
\newcommand{\red}[1]{\textcolor{red}{#1}}
\newcommand{\black}[1]{\textcolor{black}{#1}}
\newcommand{\blue}[1]{\textcolor{blue}{#1}}
\def\ie{\textit{i.e., }}
\definecolor{mygray}{gray}{.9}
\definecolor{lightblue}{RGB}{220, 230, 255} % Adjust as needed

\usepackage{tikz} % Add this in your preamble

\newcommand{\bigcircle}[1]{\tikz[baseline=(char.base)]{
    \node[shape=circle, draw, fill=black!75, text=white, inner sep=0.4pt, minimum size=4pt] (char) {\textbf{#1}};}}

\def\httilde{\mbox{\tt\raisebox{-.5ex}{\symbol{126}}}}

\begin{document}

%%%%%%%%% TITLE
\title{FaceAnonyMixer: Cancelable Faces via Identity Consistent Latent Space Mixing}

\author{Mohammed Talha Alam\textsuperscript{1}, Fahad Shamshad\textsuperscript{1}, Fakhri Karray\textsuperscript{1,2}, Karthik Nandakumar\textsuperscript{1,3} \\
\textsuperscript{1}{Mohamed Bin Zayed University of Artificial Intelligence, UAE}\\
\textsuperscript{2}{University of Waterloo, Canada} \quad  
\textsuperscript{3}{Michigan State University, USA}\\
{\tt\small \{mohammed.alam, fahad.shamshad, fakhri.karray, karthik.nandakumar\}@mbzuai.ac.ae}\\
}

\twocolumn[{%
\renewcommand\twocolumn[1][]{#1}%
\maketitle%
\vspace{-2em}%
\begin{center}
    \centering
    \includegraphics[width=\textwidth, trim = 1cm 5.6cm 0cm 3.5cm, clip]{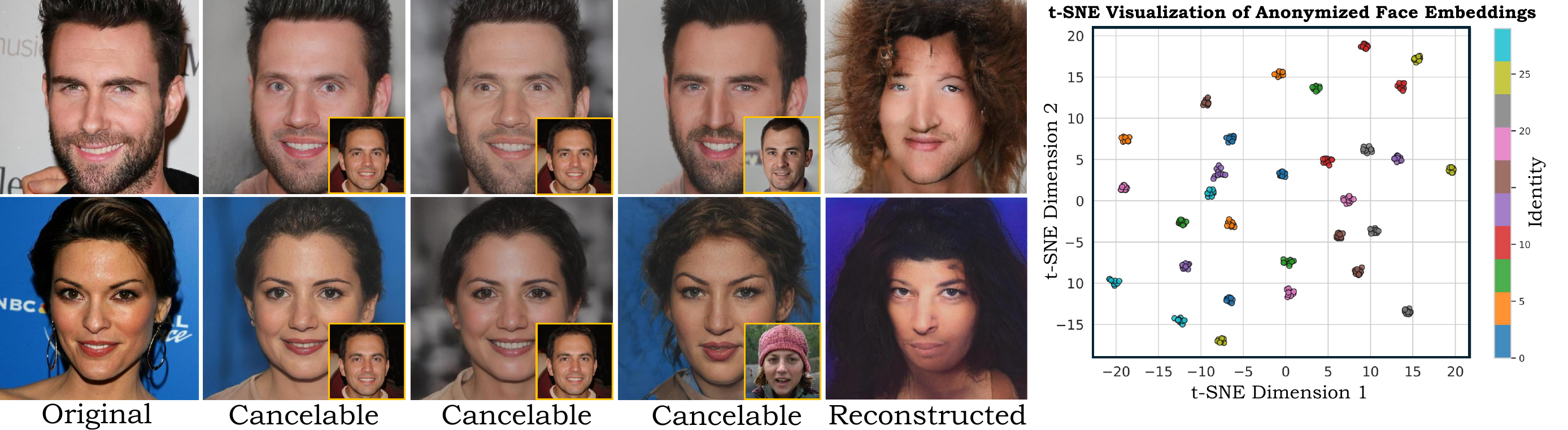}
    % \vspace{-0.3in}
    \captionof{figure}{\textbf{Visualization of our FaceAnonyMixer cancelable face generation approach}. \textit{From left to right}: \textbf{Column 1} shows original unprotected face images; \textbf{Column 2} presents protected faces generated using a fixed key (shown in the bottom-right inset), demonstrating high anonymity by visually obscuring identity. \textbf{Column 3} illustrates identity preservation, where different images of the individual, when anonymized using the same key, remain visually consistent. \textbf{Column 4} showcases revocability and unlinkability, where changing the key generates distinct protected faces, preventing cross-matching. \textbf{Column 5} highlights non-invertibility: even when both the key and protected face are known, reconstructing the original face fails.  The t-SNE plot (right) confirms identity preservation, as embeddings of protected faces from different individuals form distinct clusters in ArcFace~\cite{Deng2019Arcface} embedding space.} %This figure }
    \label{fig:introduction}
\end{center}
}]
\maketitle
\thispagestyle{empty}

%%%%%%%%% ABSTRACT
\begin{abstract}
 Advancements in face recognition (FR) technologies have amplified privacy concerns, necessitating methods that protect identity while maintaining recognition utility.
 Existing face anonymization methods typically focus on obscuring identity but fail to meet the requirements of biometric template protection, including revocability, unlinkability, and irreversibility.
 We propose \textbf{FaceAnonyMixer}, a cancelable face generation framework that leverages the latent space of a pre-trained generative model to synthesize privacy-preserving face images. The core idea of \textbf{FaceAnonyMixer} is to irreversibly mix the latent code of a real face image with a synthetic code derived from a revocable key. 
 The mixed latent code is further refined through a carefully designed multi-objective loss to satisfy all cancelable biometric requirements.
 \textbf{FaceAnonyMixer} is capable of generating high-quality cancelable faces that can be directly matched using existing FR systems without requiring any modifications.
 Extensive experiments on benchmark datasets demonstrate that \textbf{FaceAnonyMixer} delivers superior recognition accuracy while providing significantly stronger privacy protection, achieving over an 11\% absolute gain on commercial API compared to recent cancelable biometric methods.
 Code is available at: \url{https://github.com/talha-alam/faceanonymixer}
\end{abstract}

%%%%%%%%% BODY TEXT
\section{Introduction} \label{sec:intro}

Face recognition (FR) has found widespread acceptance in authentication applications due to its inherent convenience and exceptional recognition accuracy. However, unlike passwords, facial biometric templates are intrinsically linked to individuals and are irrevocable. This creates a security and privacy risk if the stored templates are compromised. The increasing prevalence of large-scale data breaches further amplifies this threat, necessitating robust privacy-preserving face biometric solutions~\cite{meden2021privacy}.

Among the various template protection methods, \textit{\textbf{cancelable face biometrics}} (CFB)~\cite{ratha2001enhancing,soutar1998biometric,patel2015cancelable,manisha2020cancelable} is a particularly promising approach. CFB applies non-invertible transformations to facial features or images to generate protected templates that maintain authentication utility while ensuring that the original identity information remains secure. These transformations, guided by user-specific keys, enable compromised templates to be revoked and replaced without requiring new biometric traits. 
To ensure both security and usability, CFB approaches must adhere to biometric protection principles outlined in ISO/IEC 24745~\cite{ISO24745:2022}:
\bigcircle{1}~\textit{\textbf{Revocability}}: A compromised template can be revoked and replaced by changing the \textit{key}, similar to password resets;
\bigcircle{2}~\textit{\textbf{Unlinkability}}: Protected templates generated from distinct keys should not be cross-matched, preventing identity tracking across different applications; 
\bigcircle{3}~\textit{\textbf{Irreversibility}}: Even if a protected template and transformation key are compromised, the original biometric data must remain irrecoverable;  
\bigcircle{4}~\textit{\textbf{Performance Preservation}}: The transformation ensures comparable recognition accuracy to unprotected biometric systems, ensuring practical utility. 

\begin{figure}[t]
    \centering
    \includegraphics[width=0.95\linewidth]{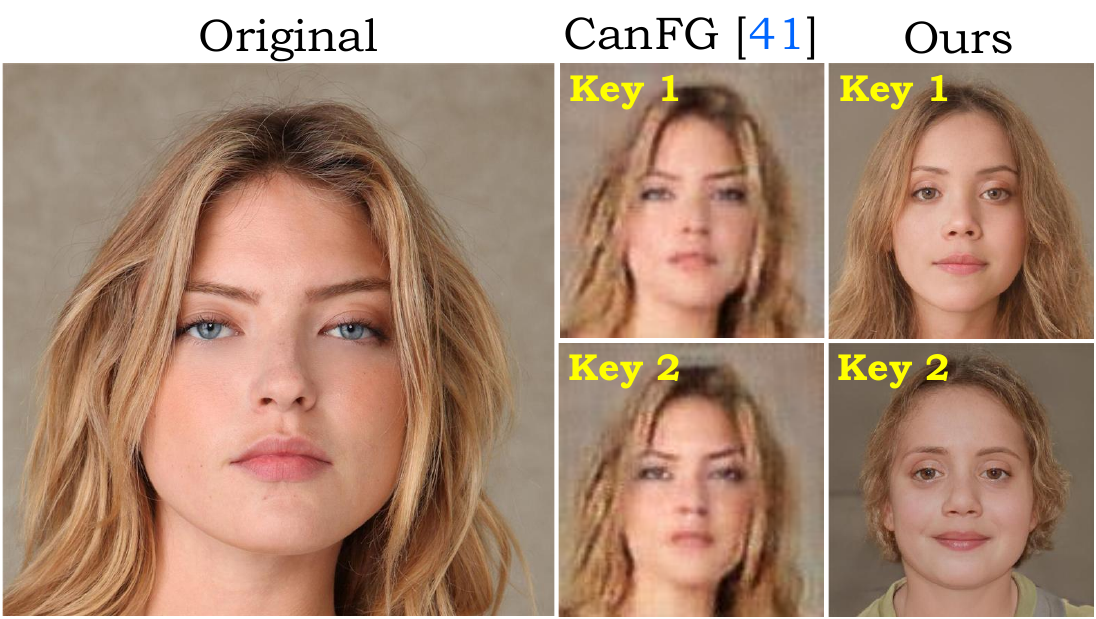}
    % \vspace{-0.7cm}
    \caption{\textbf{Comparison of protected face images generated using different keys}. \textit{Left}: Original unprotected face. \textit{Middle}: CanFG-protected faces~\cite{wang2024make} exhibit high visual similarity despite using different keys, potentially compromising \textbf{\textit{unlinkability}} and enabling cross-system tracking. Additionally, visible artifacts may degrade \textbf{\textit{recognition performance}}. \textit{Right}: Our  approach generates visually distinct, high-quality protected faces for each key, ensuring both strong \textbf{\textit{unlinkability}} and effective \textbf{\textit{revocability}} while maintaining natural appearance.}
    \vspace{-1em}
    \label{fig:diff_keys}
\end{figure}

%CFB approaches can be generally categorized into \textbf{feature-space} and \textbf{image-space} methods depending on their transformation domain~\cite{manisha2020cancelable}. Feature-space methods apply transformations to extracted facial features, which often degrade discriminative information and reduce recognition accuracy. Moreover, they typically require specialized matchers tailored to the transformed domain, limiting compatibility with standard FR pipelines. In contrast, image-space techniques operate directly on facial images before feature extraction to produce protected face images. This offers several distinct advantages: \textit{seamless integration with existing FR infrastructure}, \textit{visual interpretability of protection levels}, and \textit{preservation of utility for downstream tasks} such as facial expression analysis and age estimation.

CFB approaches can be broadly classified into \textbf{feature-space} and \textbf{image-space} methods, depending on the domain of transformation~\cite{manisha2020cancelable}. Feature-space techniques transform extracted facial embeddings; however, these transformations often degrade discriminative features, resulting in lower recognition accuracy. Furthermore, they typically require custom-designed matchers that operate in the transformed feature domain, thereby limiting their compatibility with standard FR pipelines.
In contrast, image-space methods operate directly on facial images before feature extraction to produce protected face images. This offers several key advantages: \textit{seamless integration with existing FR infrastructure}, \textit{visual interpretability of the degree of protection}, and \textit{preservation of utility for downstream tasks, such as facial expression analysis and age estimation}.

%Despite their advantages, existing image-space cancelable face schemes suffer from serious limitations.  Traditional methods such as block permutation~\cite{ratha2001enhancing} and image morphing~\cite{ghafourian2023otb} generate visually unnatural images easily flagged by modern FR systems.  Addressing these limitations, a recent deep-learning approach, Cancelable Face Generator (CanFG)~\cite{wang2024make}, generates privacy-preserving face images while supporting CFB principles. CanFG first extracts deep facial features using a pre-trained ArcFace model~\cite{Deng2019Arcface}, applies orthogonal transformations (the key) to create revocable virtual identities, and then employs a U-Net-based model to synthesize protected faces that visually differ from the original while retaining authentication utility.

Despite their benefits, current image-space CFB solutions face significant challenges. Early techniques, such as block permutation~\cite{ratha2001enhancing} and image morphing~\cite{ghafourian2023otb}, produce visually unnatural artifacts that are easily detected by modern FR systems. Building upon these efforts, the Cancelable Face Generator (CanFG)~\cite{wang2024make} offers a notable advancement, leveraging deep learning to address many of these shortcomings. CanFG extracts facial embeddings using a pre-trained ArcFace model~\cite{Deng2019Arcface}, applies orthogonal key-based transformations to create revocable identities, and employs a U-Net generator~\cite{ronneberger2015u} to synthesize protected faces.

While effective, CanFG suffers from four critical limitations. \textit{\textbf{High Retraining Cost}:} CanFG architecture necessitates end-to-end retraining from scratch for every \textit{key} alteration. This poses a significant computational bottleneck, hindering its scalability and adaptability in real-world scenarios where frequent \textit{key} updates may be necessary. 
\textbf{\textit{Insufficient Identity Obfuscation}:} The protected faces generated by CanFG often retain strong visual similarity to the original identities, as it fails to sufficiently alter identity-defining features. This compromises protection rates against advanced face recognition systems. 
\textbf{\textit{Limited Diversity}:} Although CanFG introduces diversity via orthogonal matrices, its transformation space is inherently constrained by the U-Net generator, resulting in limited variation across different protection instances of the same identity. This homogeneity increases the risk of statistical linkage attacks, compromising unlinkability (see Fig.~\ref{fig:diff_keys}). 
\textbf{\textit{Visual Quality Degradation}:} Adversarial optimization in CanFG introduces unnatural distortions in key facial regions due to the absence of explicit perceptual quality controls, as shown in Fig.~\ref{fig:diff_keys}. These artifacts reduce visual realism and can negatively impact downstream vision tasks.

To address these limitations, we propose \textbf{FaceAnonyMixer}, a novel CFB framework that operates in the latent space of a generative model, enabling the synthesis of high-quality protected faces while preserving recognition utility.
Using a \textbf{revocable key}, we sample a synthetic latent code from the generator’s latent space, where each key produces a unique, unlinkable transformation. Unlike prior methods, our approach eliminates the need for re-training when keys are changed.
Given an input face, we first perform latent inversion to obtain an identity-disentangled representation, which allows precise and controlled transformation. 
We then blend the latent code of the inverted face with that of the synthetic key face, followed by optimization using a set of carefully designed loss functions. These losses ensure the selective transfer of identity-agnostic attributes while preserving identity consistency, thus maintaining compatibility with standard FR systems. 
Our approach enforces \textbf{unlinkability} - as each key produces statistically distinct protected templates that cannot be cross-matched, and \textbf{irreversibility} - since reconstructing the original identity remains infeasible even with knowledge of key and transformation parameters.
By leveraging the structured latent representation of generative models, \textbf{FaceAnonyMixer} synthesizes visually realistic protected faces that retain \textbf{recognition performance} required for authentication.
Our main contributions are:

\begin{itemize}
    \item \textbf{FaceAnonyMixer Framework:} We introduce FaceAnonyMixer, a generative-model-based CFB framework that simultaneously satisfies the four ISO/IEC 24745 criteria — \textit{revocability}, \textit{unlinkability}, \textit{irreversibility}, and \textit{performance preservation}. Operating entirely in the latent space of a pre-trained generative model, FaceAnonyMixer ensures robust privacy protection while maintaining seamless compatibility with standard FR pipelines.
    \item \textbf{Identity Preservation Loss:} We introduce a novel \textit{identity preservation loss} that enforces identity consistency across protected face images of the same individual, ensuring reliable verification while adhering to core privacy requirements of CFB.
    \item \textbf{Comprehensive Evaluation:}  We conduct extensive experiments on multiple benchmark datasets, where \textit{FaceAnonyMixer} demonstrates superior privacy protection and recognition utility compared to existing CFB methods. We further perform comprehensive ablation studies to quantify the contribution of each module in our framework and to validate its robustness across diverse scenarios, including evaluations against widely used FR models and commercial APIs.
\end{itemize}

\section{Related Work}
\noindent \textbf{Cancelable Face Biometrics (CFB)}: 
CFB protects facial templates through non-invertible transformations to ensure revocability, unlinkability, and recognition utility~\cite{ratha2001enhancing,patel2015cancelable,bernal2023review,melzi2024overview}. Early methods such as block permutation~\cite{ratha2001enhancing} and biohashing~\cite{teoh2006random} often degrade recognition accuracy or lacked unlinkability~\cite{shahreza2023benchmarking}, while learned transformations approaches like MLP Hashing~\cite{shahreza2023mlp} improved security but remained vulnerable to template inversion attacks. Bloom filter-based approaches~\cite{rathgeb2013alignment} offered revocability but suffered from feature collisions and cross-matching risks~\cite{gomez2017general}. Image-space methods, including cartesian warping~\cite{soutar1998biometric} and face morphing~\cite{ghafourian2023otb}, improved compatibility with FR systems but introduced artifacts that limit usability. Similarly, Beyond Privacy~\cite{wang2025beyond} achieves traceability via identity transformation but does not fulfill CFB requirements like revocability and unlinkability.
Recent deep learning-based methods, such as Cancelable Face Generator (CanFG)~\cite{wang2024make}, leverage generative models to synthesize protected faces by applying orthogonal transformations to deep embeddings before image synthesis. \textit{While CanFG marks progress, it suffers from key-dependent retraining, weak identity obfuscation, and limited unlinkability, leaving it vulnerable to cross-matching. Our work overcomes these challenges by exploiting latent space manipulation in generative models to achieve unlinkable, high-quality, and revocable face transformations without retraining for key updates}.

\noindent \textbf{Generative Models for Facial Privacy}: 
Recent progress in deep generative models has led to more advanced approaches for face privacy protection~\cite{goodfellow2020generative,kingma2013auto,laishram2025toward}. Early methods like DeepPrivacy~\cite{hukkelaas2019deepprivacy} and CIAGAN~\cite{maximov2020ciagan} generated anonymized faces by replacing identity features but lacked support for revocable biometric templates, limiting their applicability to CFB. DP-CGAN~\cite{torkzadehmahani2019dp} introduced differential privacy but compromised image quality, reducing FR usability.
More recent works exploit the disentangled latent spaces of pre-trained models such as StyleGAN~\cite{karras2019style}, which enable controlled attribute editing due to their hierarchical representation~\cite{shamshad2023evading}. Methods like FaceShifter~\cite{li2019faceshifter} and InterFaceGAN~\cite{shen2020interfacegan} demonstrated how latent editing preserves realism while modifying facial attributes, inspiring privacy-preserving approaches. Subsequent methods anonymize faces in StyleGAN's latent space while preserving pose, expression, or contextual consistency~\cite{barattin2023attribute,shamshad2023clip2protect,li2023riddle}. \textit{However, most generative privacy approaches remain limited to anonymization and fail to satisfy the requirements of cancelable biometrics, including revocability, unlinkability, and non-invertibility. In contrast, our work establishes a principled latent-space framework that advances the role of generative models in CFB, delivering strong privacy guarantees while preserving recognition utility.}

\section{Proposed FaceAnonyMixer Framework}

\begin{figure*}[t]
\centering
    \includegraphics[width=0.95\textwidth, trim = 0cm 1.2cm 1cm 1cm, clip]{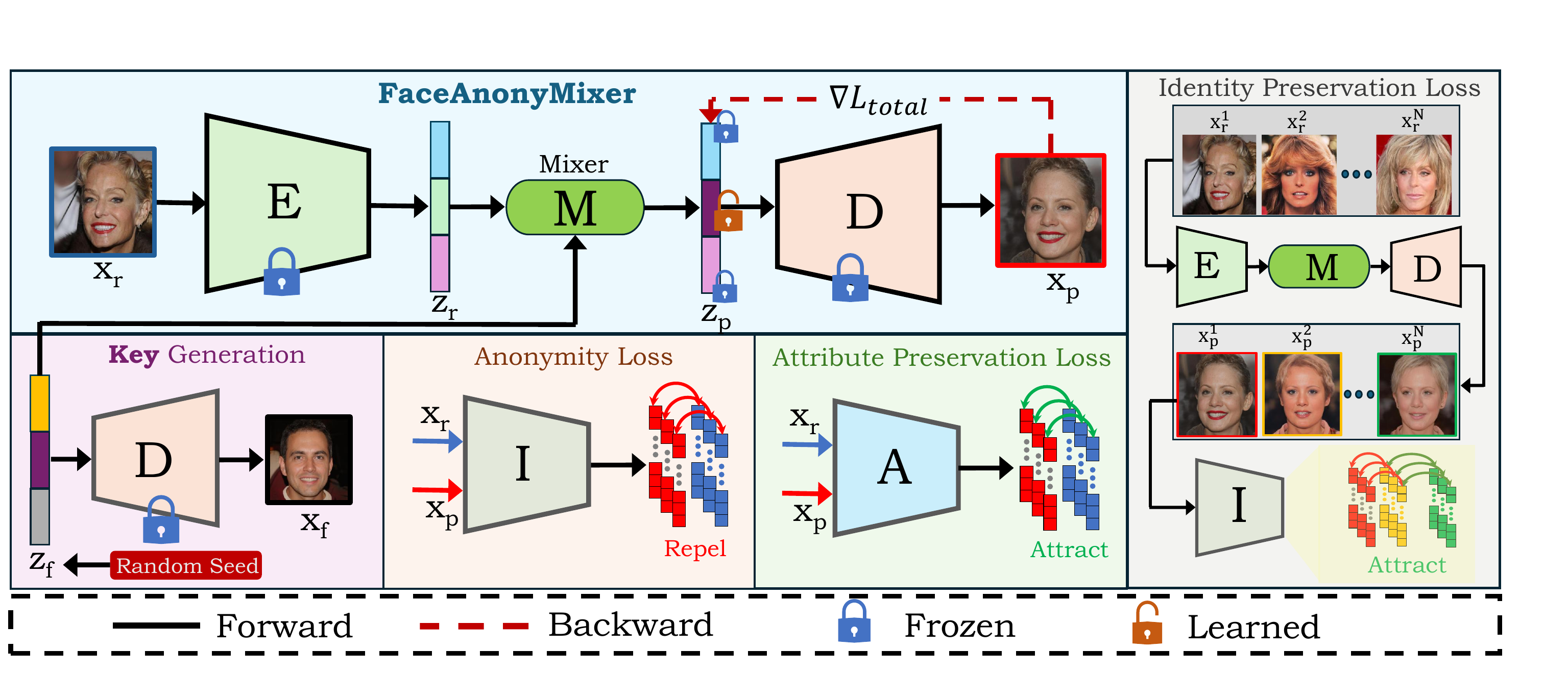}
\caption{\textbf{Overview of the FaceAnonyMixer pipeline}. A real face is first projected into StyleGAN2’s latent space, then a revocable key generates a synthetic latent code whose identity-specific layers are blended with those of the original. This hybrid latent vector is fine-tuned under three complementary losses: \textbf{anonymity loss} to suppress identity cues, \textbf{attribute-preservation loss} to keep non-identity features, and \textbf{identity-preservation loss} to retain matchability.}%Real faces are first inverted into StyleGAN2's latent space. The identity-specific latent components are replaced with a revocable ``key" (synthetic face). The resulting latent code is then optimized to ensure anonymity, attribute preservation, and identity preservation.}
\label{fig:framework}
\end{figure*}

\subsection{Background}

\textbf{Notations}: Let $\mathcal{X}$ denote the space of human face images. Let $\mathbf{F}:\mathcal{X} \times \mathcal{X} \rightarrow \{0,1\}$ denote a face matcher that outputs a binary match ($1$) or non-match ($0$) decision. Let $\mathbf{x}_1 \in \mathcal{X}$ and $\mathbf{x}_2 \in \mathcal{X}$ denote two different face images. If the two faces $\mathbf{x}_1$ and $\mathbf{x}_2$ belong to the same individual (a.k.a. identity or person), we denote it as $\mathbf{x}_1 \equiv \mathbf{x}_2$. On the other hand, $\mathbf{x}_1 \not\equiv \mathbf{x}_2$ refers to the case where $\mathbf{x}_1$ and $\mathbf{x}_2$ belong to different individuals. The face matcher is expected to output $\mathbf{F}(\mathbf{x}_1,\mathbf{x}_2) = 1$ when $\mathbf{x}_1 \equiv \mathbf{x}_2$ and $\mathbf{F}(\mathbf{x}_1,\mathbf{x}_2) = 0$ when $\mathbf{x}_1 \not\equiv \mathbf{x}_2$. Note that the face matcher may not be perfect and is typically characterized by two error rates: (i) \textit{false match rate} (FMR): $P(\mathbf{F}(\mathbf{x}_1,\mathbf{x}_2) = 1 ~|~ \mathbf{x}_1 \not\equiv \mathbf{x}_2)$, where $P$ denotes probability; and (ii) \textit{false non-match rate} (FNMR): $P(\mathbf{F}(\mathbf{x}_1,\mathbf{x}_2) = 0 ~|~ \mathbf{x}_1 \equiv \mathbf{x}_2)$.

\noindent \textbf{Problem Statement}: Our goal is to develop a cancelable face generator $\mathbf{G}: \mathcal{X} \times \mathcal{K} \rightarrow \mathcal{X}$ that takes a real face image $\mathbf{x}_r \in \mathcal{X}$ and a key $\mathbf{k} \in \mathcal{K}$ (where $\mathcal{K}$ denotes the key space) as inputs and outputs a protected face $\mathbf{x}_p \in \mathcal{X}$ such that the following four requirements are satisfied:

\begin{enumerate}
    \item \textbf{Anonymity}: The protected face image $\mathbf{x}_p$ must not match with the real face $\mathbf{x}_r$. In other words, the protection success rate defined as $P(\mathbf{F}(\mathbf{x}_r,\mathbf{x}_p) = 0 ~|~ \mathbf{x}_p=\mathbf{G}(\mathbf{x}_r,\mathbf{k}))$ must be high $~\forall~\mathbf{x}_r \in \mathcal{X} ~\text{and}~ \mathbf{k} \in \mathcal{K}$.

    \item \textbf{Identity Preservation}: Protected faces generated from different real face images of the same person using the same key must have low FNMR, i.e., if $\mathbf{x}_{p1}=\mathbf{G}(\mathbf{x}_{r1},\mathbf{k})$ and  $\mathbf{x}_{p2}=\mathbf{G}(\mathbf{x}_{r2},\mathbf{k})$, then 
    $P(\mathbf{F}(\mathbf{x}_{p1},\mathbf{x}_{p2}) = 0 ~|~ \mathbf{x}_{r1} \equiv \mathbf{x}_{r2})$ must be low. Moreover, protected faces generated from real face images of different individuals must have low FMR, even if they are generated using the same key, i.e.,  $P(\mathbf{F}(\mathbf{x}_{p1},\mathbf{x}_{p2}) = 1 ~|~ \mathbf{x}_{r1} \not\equiv \mathbf{x}_{r2})$ must be low.

    \item \textbf{Unlinkability}: Protected face images generated from the same real face using two different keys must not match, i.e., if $\mathbf{x}_{p1}=\mathbf{G}(\mathbf{x}_r,\mathbf{k}_1)$ and $\mathbf{x}_{p2}=\mathbf{G}(\mathbf{x}_r,\mathbf{k}_2)$, where $\mathbf{k}_1, \mathbf{k}_2 \in \mathcal{K},~\mathbf{k}_1 \neq \mathbf{k}_2$, then $P(\mathbf{F}(\mathbf{x}_{p1},\mathbf{x}_{p2}) = 0)$ must be high.

    \item \textbf{Irreversibility}: Suppose that it is possible to learn the inverse mapping $\mathbf{G}^{-1}: \mathcal{X} \times \mathcal{K} \rightarrow \mathcal{X}$ between the protected face and the real face using the knowledge of the key. Let $\hat{\mathbf{x}}_r$ denote the reconstructed face obtained using this inverse mapping. In this scenario, the reconstructed face $\hat{\mathbf{x}}_r$ must not match with the original face $\mathbf{x}_r$, i.e., $P(\mathbf{F}(\mathbf{x}_r,\hat{\mathbf{x}}_r) = 0 ~|~ \hat{\mathbf{x}}_r=\mathbf{G}^{-1}(\mathbf{G}(\mathbf{x}_r,\mathbf{k}),\mathbf{k}))$ must be high $~\forall~\mathbf{x}_r \in \mathcal{X} ~\text{and}~ \mathbf{k} \in \mathcal{K}$. 
\end{enumerate}

\subsection{Overview of Proposed Solution}

Our goal is to develop a cancelable face biometric framework that fulfills all ISO/IEC 24745 requirements, namely revocability, unlinkability, irreversibility, and performance preservation, while generating visually realistic protected face images. Unlike traditional anonymization methods that apply blurring~\cite{li2021deepblur} or identity replacement~\cite{li2023riddle}, our solution leverages the latent space of a pre-trained generator to achieve key-based and revocable transformations with fine-grained control over identity features.

%Our goal is to develop a privacy-preserving face anonymization framework that satisfies the requirements of cancelable biometrics. The proposed approach ensures that anonymized representations are unlinkable to the original identity, resistant to inversion attacks, and preserves identity for accurate face verification purposes. Unlike traditional anonymization methods, which either blur facial features~\cite{li2021deepblur} or replace identities with synthetic ones~\cite{li2023riddle}, our framework operates in the latent space of a pre-trained generative model to produce revocable face images.

The pipeline, illustrated in Figure~\ref{fig:framework}, first inverts the input face into the latent space of generative model using an encoder that provides an identity-disentangled representation. A revocable key is mapped to a synthetic latent code that acts as the identity-mixing counterpart. Through structured latent mixing, identity-defining components of the original face are selectively obfuscated while non-identity attributes, such as pose and expression, are preserved to maintain recognition utility and visual realism.

%As illustrated in Figure~\ref{fig:framework}, our method first encodes an input face image into the latent space of a pre-trained StyleGAN2 generator. Instead of merely perturbing the latent representation, we introduce an identity transformation mechanism that selectively obfuscates identity-related features while preserving facial attributes relevant to recognition tasks. This transformation is guided by a revocable key, ensuring that each anonymization instance remains unique yet unlinkable to prior transformations.

To achieve a balance between privacy protection and usability, the mixed latent code is optimized using a multi-objective loss function that combines anonymity, identity preservation, and attribute retention. These objectives collectively ensure unlinkability, prevent inversion attacks, and maintain compatibility with standard face recognition systems. The optimized latent code is then decoded through the generator to produce a protected face image that can be revoked and regenerated with a new key without the need for retraining, enabling scalability and adaptability.

%To balance identity preservation and revocability, our framework optimizes the transformed latent code using a multi-objective loss function. Following optimization, the modified latent code is passed through the StyleGAN2 generator to produce a protected face image that maintains a natural appearance while ensuring that it is compatible with the face matcher. Unlike prior methods that rely on fixed transformations, our approach enables dynamic revocation and regeneration of protected templates, making it well-suited for applications requiring cancelable biometrics.

% Our goal is to develop a privacy-preserving face anonymization framework that satisfies the requirements of cancelable biometrics. The proposed method operates in the latent space of a pre-trained StyleGAN2 generator to produce anonymized faces that are visually distinct from original identities while maintaining facial attributes and supporting biometric recognition functionality. As illustrated in Figure~\ref{fig:overview}, our approach first encodes a real face into StyleGAN2's latent space, then strategically replaces identity-related components with a revocable ``virtual identity key," and finally optimizes the resulting latent code to ensure non-invertibility, recognition consistency, and attribute preservation.

\subsection{Naïve FaceAnonyMixer}

Let $\mathbf{D}:\mathcal{W}^{+}\to\mathcal{X}$ be a pre-trained StyleGAN2 generator that takes an intermediate latent code $\mathbf{z}_f \in \mathcal{W}^{+}$ as input and generates a synthetic face image $\mathbf{x}_f$. Here, $\mathcal{W}^{+}$ represents the controllable latent space of StyleGAN2. Similarly, let $\mathbf{E}:\mathcal{X}\to\mathcal{W}^{+}$ be a pre-trained GAN inversion encoder that takes a real face image $\mathbf{x}_r$ as input and estimates the corresponding latent code $\mathbf{z}_r \in \mathcal{W}^{+}$. In this work, we use e4e (encoder for editing)~\cite{tov2021designing} as our encoder $\mathbf{E}$ due to its excellent trade-off between reconstruction fidelity and editability in the $\mathcal{W}^+$ space. The core idea of FaceAnonyMixer is to generate the protected (synthetic) face image $\mathbf{x}_p$ by mixing (blending) the latent code $\mathbf{z}_r$ of the given real image with a randomly sampled latent code $\mathbf{z}_f$ using the given key $\mathbf{k} \in \mathcal{K}$ as the random seed. In other words, $\mathbf{x}_p = \mathbf{D}(\mathbf{z}_p)$, where $\mathbf{z}_p = \mathbf{M}(\mathbf{z}_r,\mathbf{z}_f)$, $\mathbf{z}_r = \mathbf{E}(\mathbf{x}_r)$, and $\mathbf{z}_f = \mathbf{S}(\mathbf{k})$. Here, $\mathbf{M}:\mathcal{W}^{+} \times \mathcal{W}^{+} \to \mathcal{W}^{+}$ denotes the latent code mixing function and $\mathbf{S}:\mathcal{K} \to \mathcal{W}^{+}$ represents the function that maps a key $\mathbf{k}$ to a latent code.

\begin{figure}[t]
    \centering
        \includegraphics[width=0.48\textwidth, trim = 0cm 0.5cm 1cm 1cm, clip]{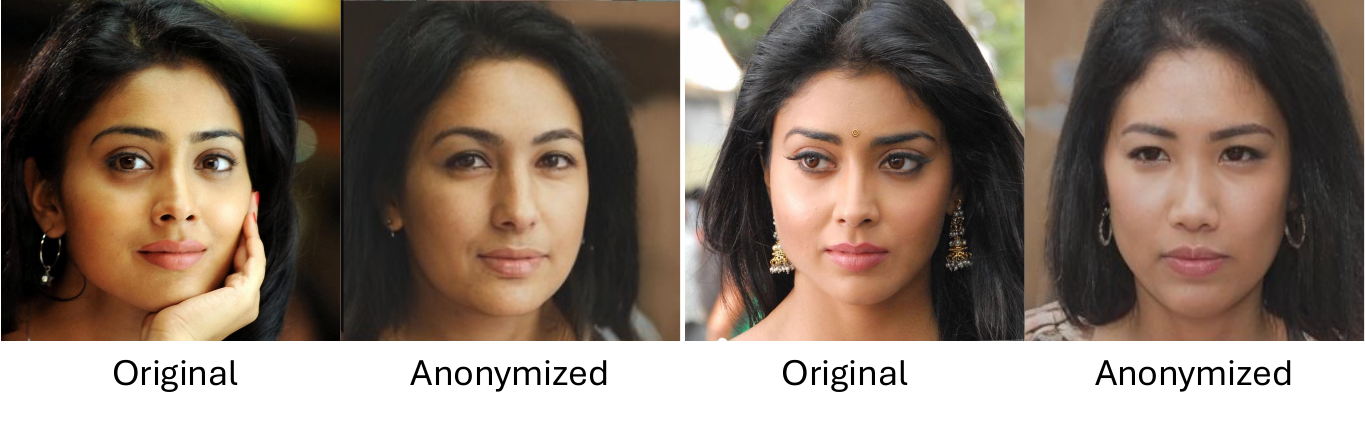}
    % \vspace{-0.7cm}
    \caption{\textbf{Limitation of the naïve FaceAnonyMixer approach}. Despite using the same key, anonymized faces (2nd and 4th columns) of the same  individual are mapped to inconsistent identities, significantly impacting recognition accuracy.}
    \vspace{-1em}
    \label{fig:naive_limit}
\end{figure}

The controllable latent space of StyleGAN2 is disentangled and typically consists of $18$ layers, where \emph{layers 0--2} control the coarse global structure (pose, face shape, etc.), \emph{layers 3--7} represent the identity-related attributes, and \emph{layers 8--17} preserve the finer details such as hair style and background~\cite{karras2019style}. Hence, an intermediate latent code $\mathbf{z} \in \mathcal{W}^{+}$ can be decomposed into three components: $\mathbf{z} = \bigl[\mathbf{z}^{(0:2)}, \;\mathbf{z}^{(3:7)}, \;\mathbf{z}^{(8:17)}\bigr]$. Consequently, a \textbf{naïve approach} to blend $\mathbf{z}_r$ and $\mathbf{z}_f$ is to replace the mid-level ``identity-related'' layers in $\mathbf{z}_r$ with the corresponding layers in $\mathbf{z}_f$, while leaving the rest of the layers untouched, i.e.,
%\vspace{-2.0em}
\begin{equation}
    \mathbf{z}_p  = \mathbf{M}_{naive} (\mathbf{z}_r,\mathbf{z}_f) =  \bigl[\mathbf{z}_r^{(0:2)}, \;\mathbf{z}_f^{(3:7)}, \;\mathbf{z}_r^{(8:17)}\bigr].
\end{equation}
The resulting protected face image $\mathbf{x}_p$ generated from $\mathbf{z}_p$ can be expected to retain the non-identity-related attributes of the given real image $\mathbf{x}_r$, while depicting the identity features of the synthetic face $\mathbf{x}_f$. Although this naïve approach disrupts the identity features and provides good anonymity, it does not preserve the identity - different face images of the same person can map to \emph{different} pseudo-identities as depicted in Fig.~\ref{fig:naive_limit} (even when the same key), thereby harming recognition performance. Furthermore, the non-identity-related attributes are not perfectly preserved by this method, leading to protected face images with poor visual quality. This phenomenon occurs due to imperfect disentanglement between the identity and non-identity attributes in the StyleGAN latent space. To address these limitations, we treat naïve mixing as a starting point and further refine the mixed latent code based on carefully designed losses.

\subsection{Optimization of FaceAnonyMixer}

To achieve all four requirements of a cancelable face generator described earlier, we further optimize the naïvely mixed latent code using a combination of four losses. Let $\mathbf{I}: \mathcal{X} \to \mathbb{R}^{d_I}$ denote a pre-trained face identity encoder such as ArcFace~\cite{Deng2019Arcface} and $\mathbf{A}: \mathcal{X} \to \mathbb{R}^{d_A}$ represent a pre-trained network for extracting identity-agnostic (non-identity-related) features such as FaRL~\cite{Zheng2022General}. Here, $d_I$ and $d_A$ represent the dimensionality of the identity and non-identity related embeddings, respectively.
\vspace{-0.5em}
\subsubsection{Anonymity Loss}
For anonymity, we must ensure the protected face is sufficiently distinct from the original face. This can be achieved by minimizing the following anonymity loss $\mathcal{L}_{\text{anon}}$ that minimizes the cosine similarity between the identity embeddings of the protected and original faces:
%\vspace{-2.0}
\begin{equation}
\label{eq:anon_loss}
\mathcal{L}_{\text{anon}} \;=\; \max\Bigl\{0,\;\cos\bigl(\mathbf{I}(\mathbf{x}_p),\mathbf{I}(\mathbf{x}_r)\bigr) - m\Bigr\},
\end{equation}

\noindent where $\mathbf{x}_p = \mathbf{D}(\mathbf{z}_p)$ and $m$ is a margin hyperparameter. Setting $m=0$ enforces maximum anonymization, pushing $\mathbf{x}_p$ towards orthogonality with $\mathbf{x}_r$ in the identity space.

\vspace{-0.5em}
\subsubsection{Identity Preservation Loss}
An identity preservation loss is \emph{critical} for cancelable biometrics because the recognition performance based on protected faces must remain comparable to that of real faces. Given a real face image $\mathbf{x}_r$, we employ an augmentation strategy to generate multiple variations (with different poses, expressions, etc.) of the same image, denoted as $\mathbf{x}_r^1,\cdots,\mathbf{x}_r^N$, where $N$ is the number of augmentations. If multiple diverse samples of the same face are already available (e.g., extracted from a video), they can also be used in lieu of the above augmentations. The following identity preservation loss $\mathcal{L}_{\text{idp}}$ attempts to ensure that the protected faces generated from multiple variations of the same real face converge in the identity embedding space:
%\vspace{-2.5}
\begin{equation}
    \label{eq:idp_loss}
    \mathcal{L}_{\text{idp}}
    = \frac{1}{N(N-1)}
    \sum_{i=1}^N
    \sum_{j=i+1}^N
    \bigl\|
    \mathbf{I}(\mathbf{x}_p^i)
    -\,
    \mathbf{I}(\mathbf{x}_p^j)
    \bigr\|_2,
\end{equation}

\noindent where $\mathbf{x}_p^i = \mathbf{D}(\mathbf{z}_p^i)$ and $\mathbf{z}_p^i = \mathbf{M}(\mathbf{E}(\mathbf{x}_r^i),\mathbf{z}_f)$.

\subsubsection{Attribute Preservation Loss}
To retain essential non-identity attributes (supporting the revocability/diversity requirement of cancelable biometrics), we minimize the following attribute preservation loss $\mathcal{L}_{\text{attr}}$:

\begin{equation}
\label{eq:attr_loss}
\mathcal{L}_{\text{attr}} = \bigl\|\mathbf{A}(\mathbf{x}_p)\;-\;\mathbf{A}(\mathbf{x}_r)\bigr\|_1.
\end{equation}

As mentioned earlier, we use a pre-trained FaRL network \cite{Zheng2022General} $\mathbf{A}$ to extract identity-agnostic features. FaRL's ViT-based encoder, trained on 20 million face image-text pairs, provides a rich semantic representation that captures facial attributes while minimizing identity information.

% \begin{figure}[t]
%     \centering
%         \includegraphics[width=0.48\textwidth, trim = 0cm 5.88cm 0cm 4cm, clip]{figures/fig_3.pdf}
%     % \vspace{-0.7cm}
%     \caption{\textbf{Limitation of the naïve FaceAnonyMixer approach}. Despite using the same key, anonymized faces (2nd and 4th columns) of the same person map to different identities, harming recognition accuracy.}
%     \label{fig:naive_limit}
% \end{figure}

\iffalse
\subsubsection{Masking Loss}
We also add a \emph{masking loss} $\mathcal{L}_{\text{mask}}$ that penalizes partial overlaps between $\mathbf{z}_p$ and the original $\mathbf{z}_r$. By penalizing direct alignment of certain latent layers, we increase the difficulty for an attacker attempting to invert $\mathbf{x}_p$ using knowledge of $\mathbf{z}_f$ (or the key $\mathbf{k}$).

\begin{equation}
\label{eq:mask_loss}
    \text{\textbf{\textcolor{red}{Need the equation for this loss}}}
\end{equation}
\fi

\begin{algorithm}[t]
\scalebox{0.93}{%
\begin{minipage}{1.0\linewidth}
%\small
\caption{FaceAnonyMixer Algorithm}
\label{alg:face-protection}

\noindent \textbf{Input:} Real face image $\mathbf{x}_r$ of a subject, a revocable key $\mathbf{k}$, StyleGAN2 generator $\mathbf{D}$, GAN inversion encoder $\mathbf{E}$, identity encoder $\mathbf{I}$, attribute encoder $\mathbf{A}$, hyperparameters $(\lambda_1,\lambda_2,\lambda_3)$, margin $m$, number of optimization steps $T$, optimization step size $\eta$, and number of augmentations $N$\\
\textbf{Output:} Protected face image $\mathbf{x}_p$

\begin{enumerate}
  \item \emph{Latent Inversion}: Invert given face image into its latent code: $\mathbf{z}_r \gets \mathbf{E}(\mathbf{x}_r)$,\quad $\mathbf{z}_r \in \mathcal{W}^{+}$.
  \item \emph{Key Mapping}: Randomly sample $\mathbf{z}_f$ from $\mathcal{W}^{+}$ using $\mathbf{k}$ as random seed: $\mathbf{z}_f \gets \mathbf{S}(\mathbf{k})$.
  \item \emph{Initialize}: Initialize $\mathbf{z}_p$ using naïve FaceAnonyMixer: $\mathbf{z}_p = \mathbf{M}_{naive} (\mathbf{z}_r,\mathbf{z}_f) =  \bigl[\mathbf{z}_r^{(0:2)}, \;\mathbf{z}_f^{(3:7)}, \;\mathbf{z}_r^{(8:17)}\bigr]$.
  \item \emph{Augment}: Generate $N$ augmentations $\mathbf{x}_r^1,\cdots,\mathbf{x}_r^N$ from $\mathbf{x}_r$, $\mathbf{z}_r^i \gets \mathbf{E}(\mathbf{x}_r^i)$, $\mathbf{z}_p^i \gets \mathbf{M}_{naive} (\mathbf{z}_r^i,\mathbf{z}_f)$, $i \in [1,N]$.
  \item \emph{Optimization Loop}:\quad\textbf{for} $t=1,\ldots,T$:
  \begin{enumerate}
    \item Generate protected faces: $\mathbf{x}_p \gets \mathbf{D}(\mathbf{z}_p)$, $\mathbf{x}_p^i \gets \mathbf{D}(\mathbf{z}_p^i)$
    \item Compute loss components $\mathcal{L}_{\text{anon}}$, $\mathcal{L}_{\text{idp}}$ and $\mathcal{L}_{\text{attr}}$ using equations (\ref{eq:anon_loss}) through (\ref{eq:attr_loss}).
    \item $\mathcal{L}_{\text{total}} = \lambda_1\,\mathcal{L}_{\text{anon}} \;+\; \lambda_2\,\mathcal{L}_{\text{idp}} \;+\; \lambda_3\,\mathcal{L}_{\text{attr}}.$
    \item Update the latent codes using gradient descent: $\mathbf{z}_p \gets \mathbf{z}_p \;-\;\eta\,\nabla_{\mathbf{z}_p}\mathcal{L}_{\text{total}}$, $\mathbf{z}_p^i \gets \mathbf{z}_p^i \;-\;\eta\,\nabla_{\mathbf{z}_p^i}\mathcal{L}_{\text{total}}$.
  \end{enumerate}
  \item \emph{Output}: Final protected face image $\mathbf{x}_p = \mathbf{D}(\mathbf{z}_p)$.
\end{enumerate}
\end{minipage}}
\end{algorithm}

\subsection{Overall FaceAnonyMixer Algorithm}
Given a real face image $\mathbf{x}_r$ and a key $\mathbf{k}$, we first apply naïve FaceAnonyMixer to obtain the initial latent code $\mathbf{z}_p$. This initial latent code is further refined by minimizing the total loss $\mathcal{L}_{\text{total}}$, which is defined as:

\begin{equation}
\mathcal{L}_{\text{total}} = \lambda_1\,\mathcal{L}_{\text{anon}} \;+\; \lambda_2\,\mathcal{L}_{\text{idp}} \;+\; \lambda_3\,\mathcal{L}_{\text{attr}}, \,%+\;\mathcal{L}_{\text{mask}},
\end{equation}

\noindent where $\lambda_1$, $\lambda_2$, and $\lambda_3$ are hyperparameters that control the relative importance of the anonymity, identity preservation, and attribute preservation losses, respectively. Let $\mathbf{z}_p^{*}$ denote the final latent code obtained after optimization. The final protected face image $\mathbf{x}_p$ can be obtained as $\mathbf{x}_p = \mathbf{D}(\mathbf{z}_p^{*})$. The overall FaceAnonyMixer algorithm is given in Alg.~\ref{alg:face-protection}.

\begin{table*}[t]
\centering
\caption{\small Protection Success Rate (\%) on CelebA-HQ and VGGFace2 against diverse FR models at different security thresholds (FMR). Higher values indicate better privacy protection, with 100\% representing perfect protection.} %where no protected faces are matched to their original counterparts.}
\label{table:tab1}
% \vspace{-2mm}
\setlength{\tabcolsep}{4pt}
\scalebox{0.9}{
\begin{tabular}{l | c || c c c c | c c c c || c }
\toprule[0.15em]
\rowcolor{mygray} \textbf{Method} & \textbf{FMR} & \multicolumn{4}{c|}{\textbf{CelebA-HQ}} & \multicolumn{4}{c||}{\textbf{VGGFace2}} & \multicolumn{1}{c}{\textbf{Average}} \\
\rowcolor{mygray} & \textbf{} & IRSE50 & IR152 & FaceNet & MobileFace & IRSE50 & IR152 & FaceNet & MobileFace &  \\
\midrule[0.15em]
 & 0.1\% & 71.85 & 70.21 & 81.48 & 34.81 & 1.09 & 15.27 & 30.68 & 0.07 & 38.18 \\
CanFG \cite{wang2024make} & 0.01\% & 60.49 & 54.40 & 71.44 & 22.88 & 0.36 & 8.54 & 21.27 & 0.05 & 29.93 \\
 & 0.001\% & 50.37 & 40.00 & 60.58 & 14.65 & 0.14 & 4.20 & 15.85 & 0.01 & 23.22 \\
\midrule
 & 0.1\% & 78.68 & 95.47 & 99.51 & 46.83 & 98.61 & 99.07 & 98.15 & 93.52 & \textbf{88.73} \\
Ours & 0.01\% & 65.58 & 84.28 & 97.28 & 29.14 & 96.76 & 97.69 & 95.37 & 85.19 & \textbf{81.41} \\
 & 0.001\% & 61.62 & 65.68 & 91.60 & 24.85 & 90.67 & 92.06 & 88.81 & 74.09 & \textbf{73.67} \\
\bottomrule[0.1em]
\end{tabular}}
\end{table*}

\begin{table*}[t]
\centering
\caption{Protection Success Rate (\%) on CelebA-HQ and VGGFace2 against diverse FR models at different FMR thresholds, measuring unlinkability. Higher PSR indicates stronger unlinkability, with 100\% meaning complete resistance to cross-matching across keys.}
\label{table:tab2}
% \vspace{-2mm}
\setlength{\tabcolsep}{4pt}
\scalebox{0.9}{
\begin{tabular}{l | c || c c c c | c c c c || c }
\toprule[0.15em]
\rowcolor{mygray} \textbf{Method} & \textbf{FMR} & \multicolumn{4}{c|}{\textbf{CelebA-HQ}} & \multicolumn{4}{c||}{\textbf{VGGFace2}} & \multicolumn{1}{c}{\textbf{Average}} \\
\rowcolor{mygray} & \textbf{} & IRSE50 & IR152 & FaceNet & MobileFace & IRSE50 & IR152 & FaceNet & MobileFace &  \\
\midrule[0.15em]
 & 0.1\% & 53.50 & 56.13 & 72.51 & 22.80 & 4.36 & 8.83 &25.18  & 2.41 & 30.72\\
CanFG~\cite{wang2024make} & 0.01\% & 40.49 & 39.59 & 58.27 & 11.28 & 2.41  & 4.63 & 16.71& 1.13 &  21.81\\
 & 0.001\% & 29.05 & 24.44 & 45.10 & 5.76 & 0.19 & 2.39& 11.8 & 0.21& 14.87\\
\midrule
 & 0.1\% & 86.21 & 95.56 & 97.58 & 47.10 &98.35 & 98.75 & 97.89 & 92.20& \textbf{89.21}  \\
Ours & 0.01\% & 66.45 & 77.82 & 95.56 & 30.97 &  95.15 & 96.14 & 95.87 & 85.21 & \textbf{80.40}\\
 & 0.001\% & 45.08 & 60.48 & 91.53 & 19.27 & 88.17 & 90.47 & 82.69 & 77.34 &\textbf{69.38} \\
\bottomrule[0.1em]
\end{tabular}}
\end{table*}
%\vspace{-1.0em}
\section{Experiments}
\noindent \textbf{Implementation Details:} 
Our method is implemented in PyTorch and evaluated on an NVIDIA A100 GPU (40GB). We use StyleGAN2 pretrained on FFHQ as the generative backbone, with latent inversion via an encoder-based approach~\cite{tov2021designing}. Identity and attribute features are extracted using ArcFace (ResNet50)~\cite{Deng2019Arcface} and FaRL (ViT-based)~\cite{Zheng2022General} encoders, respectively. Latent optimization uses Adam ($\beta_1{=}0.9$, $\beta_2{=}0.999$, $\eta{=}0.01$) for $50$ steps. Loss weights are set to $\lambda_1{=}10.0$ (identity), $\lambda_2{=}0.15$ (attribute), and $\lambda_3{=}10.0$ (consistency). For single-image identities, we generate five variations of the input image with altered poses and expressions to ensure the consistency loss is effectively applied. \\
\textbf{Datasets:} We evaluate on CelebA-HQ~\cite{karras2017progressive} and VGGFace2~\cite{cao2018vggface2}. CelebA-HQ contains 30,000 high-resolution facial images with diverse variations in pose, expression, and illumination, ideal for evaluating identity-preservation. VGGFace2 comprises 3.3 million images across 9,131 identities with substantial variations in age, pose, and background conditions, providing a challenging benchmark for assessing robustness.
Both datasets have multiple images per identity, enabling comprehensive assessment of all CFB requirements. Additional results on IJB-C~\cite{maze2018iarpa} dataset are provided in the suppl. material.\\ %anonymization effectiveness, recognition consistency, unlinkability, and non-invertibility. \\
\textbf{Face Recognition Models:} We evaluate facial privacy protection against four widely used FR models with diverse architectures in black-box settings: IRSE50~\cite{hu2018squeeze}, IR152~\cite{Deng2019Arcface}, FaceNet~\cite{schroff2015facenet}, and MobileFace~\cite{chen2018mobilefacenets}. 
All face images are aligned and cropped using MTCNN~\cite{zhang2016joint}, and performance is tested at false match rates (FMR) of 0.1\%, 0.01\%, and 0.001\%, providing a comprehensive assessment across varying security requirements. 
%Following standard protocol, all face images are aligned and cropped using MTCNN~\cite{zhang2016joint} before being processed by the FR models. We evaluate performance at multiple security thresholds corresponding to false match rates (FMR) of 0.1\%, 0.01\%, and 0.001\%, providing a comprehensive assessment across varying security requirements. 
We further assess real-world applicability against the commercial Face++ API~\cite{facepp}. \\
%Additionally, we assess privacy protection performance against the commercial Face++ API~\cite{facepp} to demonstrate real-world applicability. \\
\textbf{Evaluation Metrics:} Privacy protection is measured by the Protection Success Rate (PSR), defined as the percentage of protected faces not matched to their original identities. Higher PSR (100\% being perfect) indicates stronger protection. 
For recognition utility, we measure Equal Error Rate (EER) and Area Under the ROC Curve (AUC) when matching protected templates from the same identity. Lower EER and higher AUC indicate better recognition performance. Visual quality is assessed via Fréchet Inception Distance (FID)~\cite{heusel2017gans}, where lower scores reflect better realism.\\
\begin{table}[t]
\centering
\caption{Average confidence scores returned by the Face++ commercial API for matching protected faces to their original counterparts. Lower scores indicate better privacy protection (less similarity detected between original and protected faces).}
\label{tab:facepp_results}
% \vspace{-2mm}
\setlength{\tabcolsep}{4pt}
\scalebox{0.85}{
\begin{tabular}{l | c c c}
\toprule[0.15em]
\rowcolor{mygray} \textbf{Method} & \textbf{CelebA-HQ} & \textbf{VGGFace2} & \textbf{Average} \\
\midrule[0.15em]
CanFG~\cite{wang2024make} & 71.37 & 70.78 & 71.07 \\
Ours & \textbf{53.73} & \textbf{31.51} & \textbf{58.37} \\
\bottomrule[0.1em]
\end{tabular}}
\vspace{-1.5em}
\end{table}
\vspace{-1.0em}
\subsection{Experimental Results}
\begin{figure}[t]
\centering
\includegraphics[width=0.9\linewidth]{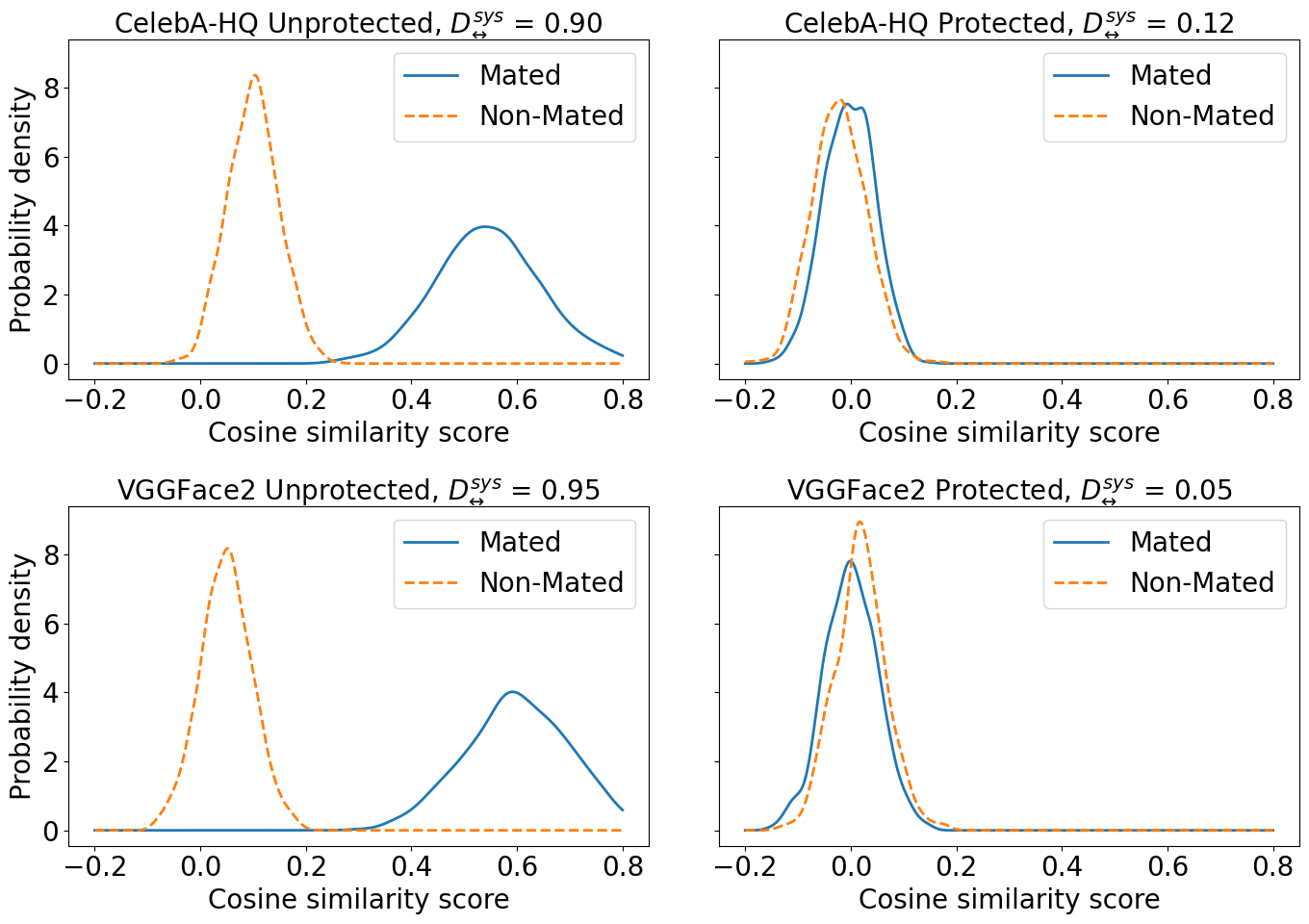}
\caption{Unlinkability analysis under Gomez-Barrero framework
on unprotected vs.\ protected templates. Solid lines are “mated” (same‑ID, different‑key) score densities; dashed lines are “non‑mated” (different‑ID) score densities.}
\label{fig:gomez}
\vspace{-1em}
\end{figure}

\begin{figure}[t]
    \centering
        \includegraphics[width=0.48\textwidth, trim = 0cm 0cm 0cm 0cm, clip]{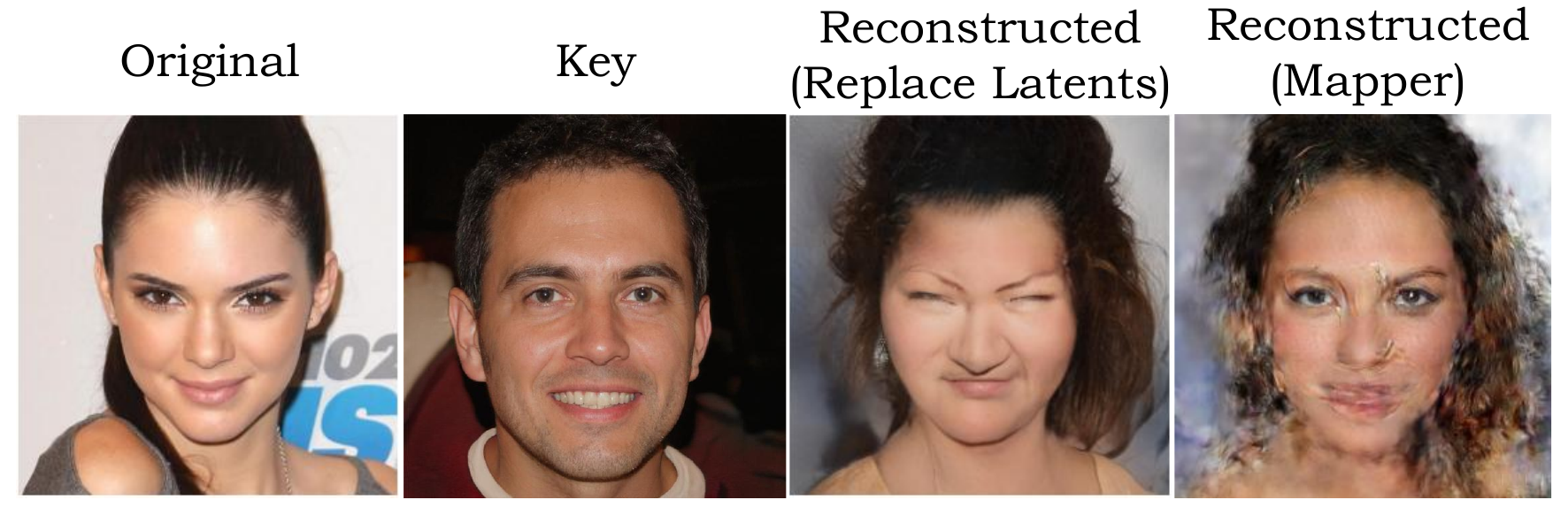}
    % \vspace{-0.7cm}
    \caption{\textbf{Non-invertibility visualization.} The original image (first) and key face (second) are used to generate reconstructed images via latent replacement (third) and a trained mapper (fourth). Both reconstructions fail to recover identity, confirming strong non-invertibility. \textit{More examples in suppl. material}.}
    \label{fig:noninvert}
\end{figure}

We evaluated our method on the held-out test sets of CelebA-HQ, consisting of 1,215 images across 307 identities following~\cite{na2022unrestricted}, and VGGFace2, which includes 1,500 images of the same identities to maintain consistency. \\
\noindent \underline{\textbf{Anonymization}}: Tab.~\ref{table:tab1} compares Protection Success Rate (PSR \%) of our method and CanFG at different False Match Rates (FMRs). Higher PSR indicates better identity protection. Our approach achieves 88.73\% average PSR at FMR = 0.1\%, significantly outperforming CanFG (38.18\%). Even at FMR = 0.001\%, our method maintains 73.67\% PSR, while CanFG drops to 23.22\%. The improvement is particularly evident on VGGFace2, where our method often exceeds 90\% PSR. These results confirm that FaceAnonyMixer provides enhanced privacy protection.\\% while preserving biometric utility. \\
\noindent \textbf{Anonymization Performance in Real-World Scenarios}: 
Tab.~\ref{tab:facepp_results} shows results against the Face++ API, where lower confidence scores indicate stronger protection.  FaceAnonyMixer achieves an average confidence score of 58.37, significantly outperforming CanFG (71.07). The gap is most pronounced on VGGFace2 (31.51 vs. 70.78), highlighting our method’s effectiveness on diverse, unconstrained images. These results demonstrate that our generative latent space approach provides superior protection against commercial recognition systems, consistently preventing matches between protected and original faces. \\
\noindent \underline{\textbf{Unlinkability}}: We assess unlinkability by measuring the Protection Success Rate (PSR) when the same face is anonymized with different keys (Tab.~\ref{table:tab2}). A higher PSR indicates that transformed identities remain distinct and resistant to cross-matching across applications. Our method achieves an average PSR of 89.21\% at FMR = 0.1\%, significantly outperforming CanFG (30.72\%). Even at FMR = 0.001\%, it maintains 69.38\%, while CanFG drops to 14.87\%, revealing its susceptibility to identity linkage. To complement PSR, we also evaluate unlinkability using the Gomez-Barrero framework~\cite{gomez2017general}. As shown in Fig.~\ref{fig:gomez}, the global unlinkability score 
$D^{sys}_{\leftrightarrow}$ on CelebA-HQ drops from $0.90$ (raw templates) to $0.12$ (protected templates), indicating a substantial reduction in cross-matching risk. Similar trends are observed on VGGFace2, further confirming that our approach ensures robust unlinkability. \\
%In Tab.~\ref{table:tab2}, we evaluate unlinkability by measuring the PSR when the same face is anonymized using different keys. A higher PSR indicates that the transformed identities remain distinct and unmatchable, preventing cross-matching across different applications. Our method achieves stronger unlinkability, with an 89.21\% average PSR at FMR = 0.1\%, far surpassing CanFG (30.72\%). Even at FMR = 0.001\%, it maintains 69.38\% PSR, while CanFG drops to 14.87\%, revealing its susceptibility to identity linkage. This validates that our approach ensures robust unlinkability, making it highly resistant to re-identification attempts across different anonymization instances.\\ 
\noindent \hspace{-1em}\underline{\textbf{Non-Invertibility}}: For non-invertibility, we assume an attacker has both the anonymized image and the latent code of the fake key face. The attacker inverts the anonymized image into StyleGAN’s latent space and attempts reconstruction by replacing the middle layers of the anonymized latent code with those of the fake key face. As shown in Fig.~\ref{fig:noninvert}, the resulting images show no resemblance to the original, confirming that direct inversion fails.
We further consider a stronger attack where the attacker collects a dataset of anonymized–original pairs to train an image-to-image translation model~\cite{isola2017image}. As Tab.~\ref{table:noninvert} shows, the high PSR values across all FR models confirm that even with a learned mapping, identity recovery remains infeasible.\\% (also see Fig.~\ref{fig:noninvert}).\\ %Our approach achieves PSR $>$ 97\% at FMR = 0.1\%, demonstrating strong non-invertibility across both datasets.
% \begin{figure}[h]
%     \centering
%     \begin{subfigure}{0.48\textwidth}
%         \centering
%         \includegraphics[width=\textwidth]{similarity_scores_1.png}
%         \caption{Match vs Non-Match Scores (Same Key)}
%         \label{fig:similarity1}
%     \end{subfigure}
%     \hfill
%     \begin{subfigure}{0.48\textwidth}
%         \centering
%         \includegraphics[width=\textwidth]{similarity_scores_2.png}
%         \caption{Match vs Non-Match Scores (Different Key)}
%         \label{fig:similarity2}
%     \end{subfigure}
%     \caption{Comparison of Similarity Score Distributions for Anonymized Data}
%     \label{fig:similarity_comparison}
% \end{figure}
\noindent \textbf{Performance Preservation}: 
Tab.~\ref{tab:rec_per} reports post-anonymization recognition performance using EER (lower is better) and AUC (higher is better). Our method consistently outperforms CanFG, achieving lower EER (0.019 vs. 0.045) and higher AUC (0.997 vs. 0.988) on CelebA-HQ, with similar gains on VGGFace2. Additionally, our approach yields higher-quality anonymized images, as reflected by a substantially lower FID of 37.73 compared to CanFG’s 82.52. These results demonstrate that our method preserves authentication performance while ensuring superior visual fidelity.

\begin{table}[t]
\centering
\caption{Protection Success Rate (PSR \%) for non-invertibility on CelebA-HQ and VGGFace2 at different security thresholds (FMR). Higher PSR values indicate stronger resistance to inversion attacks, ensuring that an adversary cannot reconstruct the original identity from anonymized faces.}
\label{table:noninvert}
% \vspace{-2mm}
\setlength{\tabcolsep}{4pt}
\scalebox{0.8}{
\begin{tabular}{l | c || c c c c}
\toprule[0.15em]
\rowcolor{mygray} \textbf{Dataset} & \textbf{FMR} & \multicolumn{4}{c}{\textbf{Non-Invertibility}} \\
\rowcolor{mygray} & \textbf{} & IRSE50 & IR152 & FaceNet & MobileFace \\
\midrule[0.15em]
 & 0.1\% & 98.02 & 98.77 & 99.26 & 96.79 \\
CelebA-HQ & 0.01\% & 94.57 & 97.53 & 96.79 & 88.97 \\
 & 0.001\% & 87.74 & 92.02 & 93.50 & 74.24 \\
\midrule
 & 0.1\% & 97.60 & 98.40 & 99.15 & 93.80 \\
VGGFace2 & 0.01\% & 94.18 & 97.01 & 95.12 & 86.57 \\
 & 0.001\% & 91.35 & 95.27 & 93.8 & 75.98 \\
\bottomrule[0.1em]
\end{tabular}}
\end{table}

% In preamble:
% \usepackage{sidecap}

\begin{SCtable}[4][t]
\renewcommand{\arraystretch}{1.0}
\setlength{\tabcolsep}{2.7pt}
\scalebox{0.82}{
\begin{tabular}{lcccc}
\toprule
\rowcolor{mygray} & \multicolumn{2}{c}{\textbf{CelebA-HQ}} & \multicolumn{2}{c}{\textbf{VGGFace2}} \\
\rowcolor{mygray} Method & EER$\downarrow$ & AUC$\uparrow$ & EER$\downarrow$ & AUC$\uparrow$ \\
\midrule
Original & 0.027 & 0.990 & 0.074 & 0.964 \\
CanFG~\cite{wang2024make}& 0.045 & 0.988 & 0.101 & 0.951 \\
\textbf{Ours} & \textbf{0.019} & \textbf{0.997} & \textbf{0.059} & \textbf{0.975} \\
\bottomrule
\end{tabular}}
\captionsetup{justification=raggedright, singlelinecheck=false} % Fix spacing issue
\caption{Recognition performance of different anonymization methods.\\}
\label{tab:rec_per}
\vspace{-0.5em}
\end{SCtable}

\subsection{Ablation Studies}
\begin{figure}[t]
    \centering
    \includegraphics[width=0.46\textwidth, trim = 0cm 0cm 0cm 0cm, clip]{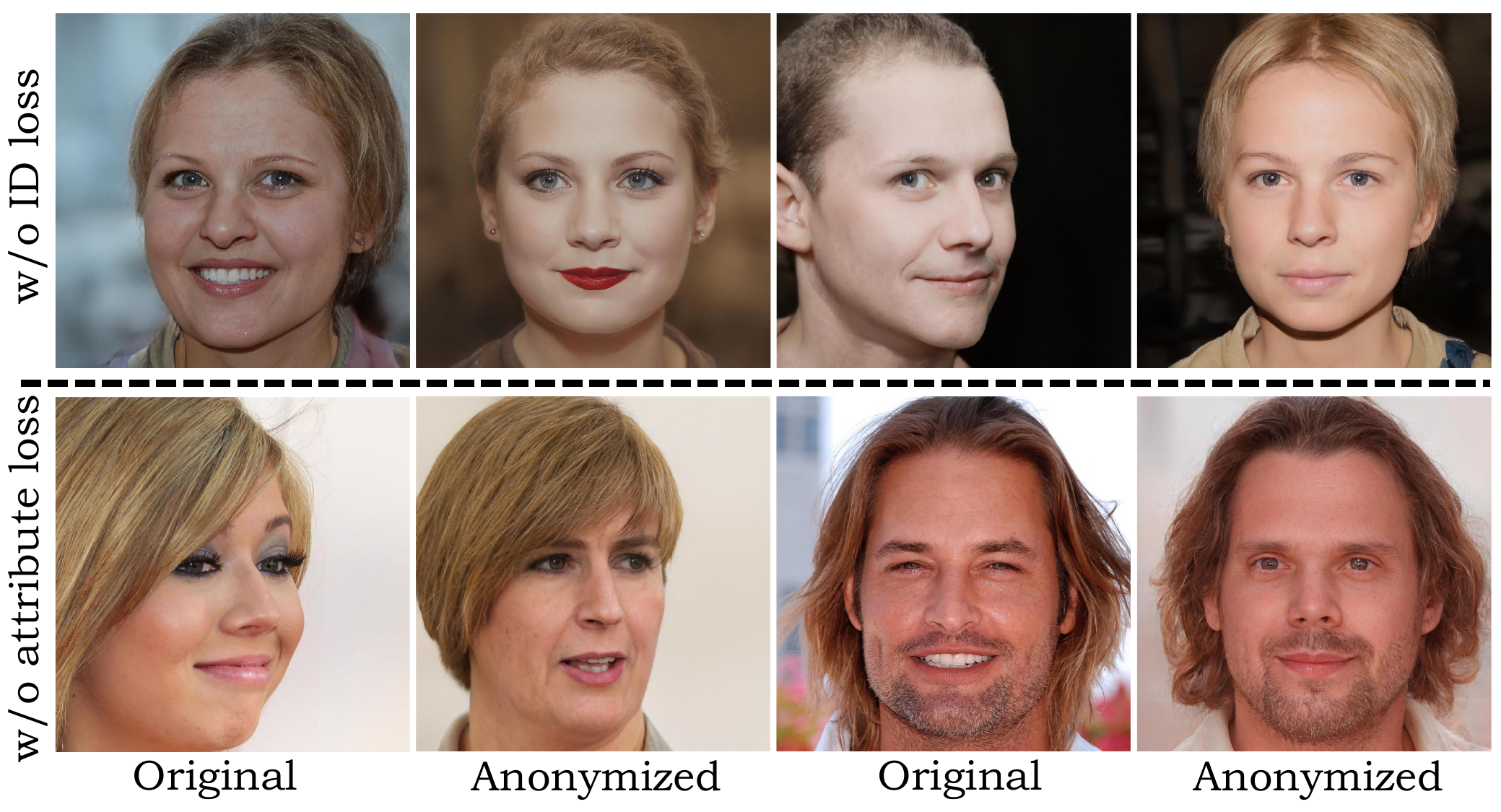}
    % \vspace{-0.7cm}
    \caption{\textbf{Ablation study on identity and attribute-preserving losses}. The top row (w/o identity loss) shows inconsistent identities despite the same key, reducing authentication reliability. The bottom row (w/o attribute loss) fails to retain facial attributes like smiles, affecting realism. \textit{For more results see suppl. material}.}
    \label{fig:abl}
\end{figure}

\noindent \textbf{Effect of Identity-Preserving Loss}: Removing the identity-preserving loss leads to highly inconsistent transformed identities, even when the same key is applied, which significantly degrades authentication performance as shown in Fig.~\ref{fig:abl}. This further highlights the critical role of this loss in maintaining identity consistency.\\
%Without the identity-preserving loss, transformed identities appear inconsistent, even when using the same key, reducing authentication performance (see Fig.~\ref{fig:abl}). This demonstrates the importance of ensuring identity consistency.\\
\noindent \textbf{Effect of Attribute-Preserving Loss}: As illustrated in Fig.~\ref{fig:abl}, excluding the attribute-preserving loss results in the loss of key facial attributes, such as expressions, after anonymization. This underlines the importance of this loss for retaining non-identity attributes, enhancing both visual realism and usability for downstream tasks.\\
%As shown in Fig.~\ref{fig:abl}, without the attribute-preserving loss, facial attributes such as expression are not retained post-anonymization. This confirms the necessity of this loss for preserving non-identity attributes, ensuring more realism and usability.\\
\noindent \textbf{Robustness Against Key Variations:}  Our method exhibits minimal variation in PSR, with an average of 93.51\% and a standard deviation of 2.33 as shown in Tab.~\ref{tab:key_robustness}, demonstrating high stability across different keys. This confirms that our framework maintains robust unlinkability and consistent performance regardless of key selection.
%The results demonstrate minimal variation in PSR, with an average score of 93.51\% and a standard deviation of 2.33 as shown in Tab. \ref{tab:key_robustness}, indicating that our method remains highly stable regardless of the key used. This confirms that our anonymization framework does not introduce inconsistencies due to key selection, ensuring robust unlinkability and performance consistency across varying conditions.
\begin{table}[t]
    \centering
    \renewcommand{\arraystretch}{1.2}
    \setlength{\tabcolsep}{9pt}
    \caption{Impact of different encryption keys on PSR (via IR152). Five different random keys were used to evaluate the robustness of our method. Std. denotes standard deviation.}
    \label{tab:key_robustness}
    \scalebox{0.8}{ % Adjusts the table size if needed
    \begin{tabular}{c|ccccc|c}
        \toprule
        \rowcolor{gray!20} & $key_1$ & $key_2$ & $key_3$ & $key_4$ & $key_5$ & Std. \\
        \midrule
        PSR & 95.47 & 90.47 & 91.91& 96.84 &  92.87& 2.33 \\
        \bottomrule
    \end{tabular}
    }
    \vspace{-1em}
\end{table}

\section{Conclusion}
We proposed FaceAnonyMixer, a novel cancelable face biometric framework that leverages the latent space of a generative model to provide robust privacy protection. By strategically blending the latent representations of original faces with synthetic key-driven codes, our method fully satisfies all core cancelable biometrics requirements, including revocability, unlinkability, irreversibility, and identity preservation. Extensive experiments on multiple benchmark datasets demonstrate that FaceAnonyMixer gives superior privacy protection against widely used FR models and commercial APIs, while maintaining verification accuracy comparable to unprotected baselines.
Future work will focus on enhancing computational efficiency and extending the proposed framework for multi-biometric systems.

{\small
\bibliographystyle{ieee}
\bibliography{egbib}

\begin{thebibliography}{10}\itemsep=-1pt

\bibitem{barattin2023attribute}
S.~Barattin, C.~Tzelepis, I.~Patras, and N.~Sebe.
\newblock Attribute-preserving face dataset anonymization via latent code optimization.
\newblock In {\em Proceedings of the IEEE/CVF conference on computer vision and pattern recognition}, pages 8001--8010, 2023.

\bibitem{bernal2023review}
J.~C. Bernal-Romero, J.~M. Ramirez-Cortes, J.~D.~J. Rangel-Magdaleno, P.~Gomez-Gil, H.~Peregrina-Barreto, and I.~Cruz-Vega.
\newblock A review on protection and cancelable techniques in biometric systems.
\newblock {\em Ieee Access}, 11:8531--8568, 2023.

\bibitem{cao2018vggface2}
Q.~Cao, L.~Shen, W.~Xie, O.~M. Parkhi, and A.~Zisserman.
\newblock Vggface2: A dataset for recognising faces across pose and age.
\newblock In {\em 2018 13th IEEE international conference on automatic face \& gesture recognition (FG 2018)}, pages 67--74. IEEE, 2018.

\bibitem{chen2018mobilefacenets}
S.~Chen, Y.~Liu, X.~Gao, and Z.~Han.
\newblock Mobilefacenets: Efficient cnns for accurate real-time face verification on mobile devices.
\newblock In {\em Chinese conference on biometric recognition}, pages 428--438. Springer, 2018.

\bibitem{Deng2019Arcface}
J.~Deng, J.~Guo, N.~Xue, and S.~Zafeiriou.
\newblock Arcface: Additive angular margin loss for deep face recognition.
\newblock In {\em Proceedings of the IEEE/CVF conference on computer vision and pattern recognition}, pages 4690--4699, 2019.

\bibitem{ghafourian2023otb}
M.~Ghafourian, J.~Fierrez, R.~Vera-Rodriguez, A.~Morales, and I.~Serna.
\newblock Otb-morph: One-time biometrics via morphing.
\newblock {\em Machine Intelligence Research}, 20(6):855--871, 2023.

\bibitem{gomez2017general}
M.~Gomez-Barrero, J.~Galbally, C.~Rathgeb, and C.~Busch.
\newblock General framework to evaluate unlinkability in biometric template protection systems.
\newblock {\em IEEE Transactions on Information Forensics and Security}, 13(6):1406--1420, 2017.

\bibitem{goodfellow2020generative}
I.~Goodfellow, J.~Pouget-Abadie, M.~Mirza, B.~Xu, D.~Warde-Farley, S.~Ozair, A.~Courville, and Y.~Bengio.
\newblock Generative adversarial networks.
\newblock {\em Communications of the ACM}, 63(11):139--144, 2020.

\bibitem{heusel2017gans}
M.~Heusel, H.~Ramsauer, T.~Unterthiner, B.~Nessler, and S.~Hochreiter.
\newblock Gans trained by a two time-scale update rule converge to a local nash equilibrium.
\newblock {\em Advances in neural information processing systems}, 30, 2017.

\bibitem{hu2018squeeze}
J.~Hu, L.~Shen, and G.~Sun.
\newblock Squeeze-and-excitation networks.
\newblock In {\em Proceedings of the IEEE conference on computer vision and pattern recognition}, pages 7132--7141, 2018.

\bibitem{hukkelaas2019deepprivacy}
H.~Hukkel{\aa}s, R.~Mester, and F.~Lindseth.
\newblock Deepprivacy: A generative adversarial network for face anonymization.
\newblock In {\em International symposium on visual computing}, pages 565--578. Springer, 2019.

\bibitem{ISO24745:2022}
{ISO/IEC JTC1 SC27 Security Techniques}.
\newblock {Information Technology - Security Techniques - Biometric Information Protection}, 2022.

\bibitem{isola2017image}
P.~Isola, J.-Y. Zhu, T.~Zhou, and A.~A. Efros.
\newblock Image-to-image translation with conditional adversarial networks.
\newblock In {\em Proceedings of the IEEE conference on computer vision and pattern recognition}, pages 1125--1134, 2017.

\bibitem{karras2017progressive}
T.~Karras, T.~Aila, S.~Laine, and J.~Lehtinen.
\newblock Progressive growing of gans for improved quality, stability, and variation.
\newblock {\em arXiv preprint arXiv:1710.10196}, 2017.

\bibitem{karras2019style}
T.~Karras, S.~Laine, and T.~Aila.
\newblock A style-based generator architecture for generative adversarial networks.
\newblock In {\em Proceedings of the IEEE/CVF conference on computer vision and pattern recognition}, pages 4401--4410, 2019.

\bibitem{kingma2013auto}
D.~P. Kingma, M.~Welling, et~al.
\newblock Auto-encoding variational bayes.
\newblock 2013.

\bibitem{laishram2025toward}
L.~Laishram, M.~Shaheryar, J.~T. Lee, and S.~K. Jung.
\newblock Toward a privacy-preserving face recognition system: A survey of leakages and solutions.
\newblock {\em ACM Computing Surveys}, 57(6):1--38, 2025.

\bibitem{li2023riddle}
D.~Li, W.~Wang, K.~Zhao, J.~Dong, and T.~Tan.
\newblock Riddle: Reversible and diversified de-identification with latent encryptor.
\newblock {\em arXiv preprint arXiv:2303.05171}, 2023.

\bibitem{li2019faceshifter}
L.~Li, J.~Bao, H.~Yang, D.~Chen, and F.~Wen.
\newblock Faceshifter: Towards high fidelity and occlusion aware face swapping.
\newblock {\em arXiv preprint arXiv:1912.13457}, 2019.

\bibitem{li2021deepblur}
T.~Li and M.~S. Choi.
\newblock Deepblur: A simple and effective method for natural image obfuscation.
\newblock {\em arXiv preprint arXiv:2104.02655}, 1:3, 2021.

\bibitem{manisha2020cancelable}
Manisha and N.~Kumar.
\newblock Cancelable biometrics: a comprehensive survey.
\newblock {\em Artificial Intelligence Review}, 53(5):3403--3446, 2020.

\bibitem{maximov2020ciagan}
M.~Maximov, I.~Elezi, and L.~Leal-Taix{\'e}.
\newblock Ciagan: Conditional identity anonymization generative adversarial networks.
\newblock In {\em Proceedings of the IEEE/CVF conference on computer vision and pattern recognition}, pages 5447--5456, 2020.

\bibitem{maze2018iarpa}
B.~Maze, J.~Adams, J.~A. Duncan, N.~Kalka, T.~Miller, C.~Otto, A.~K. Jain, W.~T. Niggel, J.~Anderson, J.~Cheney, et~al.
\newblock Iarpa janus benchmark-c: Face dataset and protocol.
\newblock In {\em 2018 international conference on biometrics (ICB)}, pages 158--165. IEEE, 2018.

\bibitem{meden2021privacy}
B.~Meden, P.~Rot, P.~Terh{\"o}rst, N.~Damer, A.~Kuijper, W.~J. Scheirer, A.~Ross, P.~Peer, and V.~{\v{S}}truc.
\newblock Privacy--enhancing face biometrics: A comprehensive survey.
\newblock {\em IEEE Transactions on Information Forensics and Security}, 16:4147--4183, 2021.

\bibitem{facepp}
{MEGVII Technology}.
\newblock Face++ cognitive services.
\newblock \url{https://www.faceplusplus.com/}, 2024.
\newblock Accessed: 2024-03-08.

\bibitem{melzi2024overview}
P.~Melzi, C.~Rathgeb, R.~Tolosana, R.~Vera-Rodriguez, and C.~Busch.
\newblock An overview of privacy-enhancing technologies in biometric recognition.
\newblock {\em ACM Computing Surveys}, 56(12):1--28, 2024.

\bibitem{na2022unrestricted}
D.~Na, S.~Ji, and J.~Kim.
\newblock Unrestricted black-box adversarial attack using gan with limited queries.
\newblock In {\em European Conference on Computer Vision}, pages 467--482. Springer, 2022.

\bibitem{patel2015cancelable}
V.~M. Patel, N.~K. Ratha, and R.~Chellappa.
\newblock Cancelable biometrics: A review.
\newblock {\em IEEE signal processing magazine}, 32(5):54--65, 2015.

\bibitem{ratha2001enhancing}
N.~K. Ratha, J.~H. Connell, and R.~M. Bolle.
\newblock Enhancing security and privacy in biometrics-based authentication systems.
\newblock {\em IBM systems Journal}, 40(3):614--634, 2001.

\bibitem{rathgeb2013alignment}
C.~Rathgeb, F.~Breitinger, and C.~Busch.
\newblock Alignment-free cancelable iris biometric templates based on adaptive bloom filters.
\newblock In {\em 2013 international conference on biometrics (ICB)}, pages 1--8. IEEE, 2013.

\bibitem{ronneberger2015u}
O.~Ronneberger, P.~Fischer, and T.~Brox.
\newblock U-net: Convolutional networks for biomedical image segmentation.
\newblock In {\em International Conference on Medical image computing and computer-assisted intervention}, pages 234--241. Springer, 2015.

\bibitem{schroff2015facenet}
F.~Schroff, D.~Kalenichenko, and J.~Philbin.
\newblock Facenet: A unified embedding for face recognition and clustering.
\newblock In {\em Proceedings of the IEEE conference on computer vision and pattern recognition}, pages 815--823, 2015.

\bibitem{shahreza2023mlp}
H.~O. Shahreza, V.~K. Hahn, and S.~Marcel.
\newblock Mlp-hash: Protecting face templates via hashing of randomized multi-layer perceptron.
\newblock In {\em 2023 31st European Signal Processing Conference (EUSIPCO)}, pages 605--609. IEEE, 2023.

\bibitem{shahreza2023benchmarking}
H.~O. Shahreza, P.~Melzi, D.~Osorio-Roig, C.~Rathgeb, C.~Busch, S.~Marcel, R.~Tolosana, and R.~Vera-Rodriguez.
\newblock Benchmarking of cancelable biometrics for deep templates.
\newblock {\em arXiv preprint arXiv:2302.13286}, 2023.

\bibitem{shamshad2023clip2protect}
F.~Shamshad, M.~Naseer, and K.~Nandakumar.
\newblock Clip2protect: Protecting facial privacy using text-guided makeup via adversarial latent search.
\newblock In {\em Proceedings of the IEEE/CVF Conference on Computer Vision and Pattern Recognition}, pages 20595--20605, 2023.

\bibitem{shamshad2023evading}
F.~Shamshad, K.~Srivatsan, and K.~Nandakumar.
\newblock Evading forensic classifiers with attribute-conditioned adversarial faces.
\newblock In {\em Proceedings of the IEEE/CVF Conference on Computer Vision and Pattern Recognition}, pages 16469--16478, 2023.

\bibitem{shen2020interfacegan}
Y.~Shen, C.~Yang, X.~Tang, and B.~Zhou.
\newblock Interfacegan: Interpreting the disentangled face representation learned by gans.
\newblock {\em IEEE transactions on pattern analysis and machine intelligence}, 44(4):2004--2018, 2020.

\bibitem{soutar1998biometric}
C.~Soutar, D.~Roberge, A.~Stoianov, R.~Gilroy, and B.~V. Kumar.
\newblock Biometric encryption using image processing.
\newblock In {\em Optical Security and Counterfeit Deterrence Techniques II}, volume 3314, pages 178--188. SPIE, 1998.

\bibitem{teoh2006random}
A.~B. Teoh, A.~Goh, and D.~C. Ngo.
\newblock Random multispace quantization as an analytic mechanism for biohashing of biometric and random identity inputs.
\newblock {\em IEEE transactions on pattern analysis and machine intelligence}, 28(12):1892--1901, 2006.

\bibitem{torkzadehmahani2019dp}
R.~Torkzadehmahani, P.~Kairouz, and B.~Paten.
\newblock Dp-cgan: Differentially private synthetic data and label generation.
\newblock In {\em Proceedings of the IEEE/CVF Conference on Computer Vision and Pattern Recognition Workshops}, pages 0--0, 2019.

\bibitem{tov2021designing}
O.~Tov, Y.~Alaluf, Y.~Nitzan, O.~Patashnik, and D.~Cohen-Or.
\newblock Designing an encoder for stylegan image manipulation.
\newblock {\em ACM Transactions on Graphics (TOG)}, 40(4):1--14, 2021.

\bibitem{wang2025beyond}
T.~Wang, W.~Wen, X.~Xiao, Z.~Hua, Y.~Zhang, and Y.~Fang.
\newblock Beyond privacy: Generating privacy-preserving faces supporting robust image authentication.
\newblock {\em IEEE Transactions on Information Forensics and Security}, 2025.

\bibitem{wang2024make}
T.~Wang, Y.~Zhang, X.~Xiao, L.~Yuan, Z.~Xia, and J.~Weng.
\newblock Make privacy renewable! generating privacy-preserving faces supporting cancelable biometric recognition.
\newblock In {\em Proceedings of the 32nd ACM International Conference on Multimedia}, pages 10268--10276, 2024.

\bibitem{zhang2016joint}
K.~Zhang, Z.~Zhang, Z.~Li, and Y.~Qiao.
\newblock Joint face detection and alignment using multitask cascaded convolutional networks.
\newblock {\em IEEE signal processing letters}, 23(10):1499--1503, 2016.

\bibitem{Zheng2022General}
Y.~Zheng, H.~Yang, T.~Zhang, J.~Bao, D.~Chen, Y.~Huang, L.~Yuan, D.~Chen, M.~Zeng, and F.~Wen.
\newblock General facial representation learning in a visual-linguistic manner.
\newblock In {\em Proceedings of the IEEE/CVF conference on computer vision and pattern recognition}, pages 18697--18709, 2022.

\end{thebibliography}
}

\twocolumn[
\begin{@twocolumnfalse}
\section*{\centering \Large{{Supplementary Material} -- FaceAnonyMixer: Cancelable Faces via Identity Consistent Latent Space Mixing}}
\end{@twocolumnfalse}
]

\vspace{5em}
This supplementary material provides additional experiments and visualizations that reinforce the effectiveness of \textit{FaceAnonyMixer} across all core requirements of cancelable biometrics. Specifically, we include:
\begin{itemize}
    \item Extended quantitative evaluation of recognition performance and post-anonymization identity separability.
    \item Embedding space visualizations using t-SNE to illustrate identity preservation and inter-class separation.
    \item Ablation studies that isolate the contribution of attribute loss, identity-consistency loss, and anonymity loss loss term.
    \item Inversion resistance experiments demonstrating robustness against reconstruction attacks.
    \item Qualitative multi-key results showcasing unlinkability and attribute retention.
    \item Qualitative results of FaceAnonyMixer on the IJB-C dataset [{\color{blue}23}].
\end{itemize}

%This supplementary material provides additional evidence supporting the effectiveness of \textit{FaceAnonyMixer} across core cancelable biometric criteria. Specifically, we present:  
%(1) quantitative analysis of recognition performance and identity separability after anonymization,  
%(2) embedding space visualizations using t-SNE,  
%(3) results from inversion resistance experiments,  
%(4) targeted ablation studies to isolate the contribution of each loss term, and   (5) qualitative multi-key results showing unlinkability and attribute preservation.\\  
\begin{figure}[htp]
\centering
\includegraphics[width=0.7\linewidth]{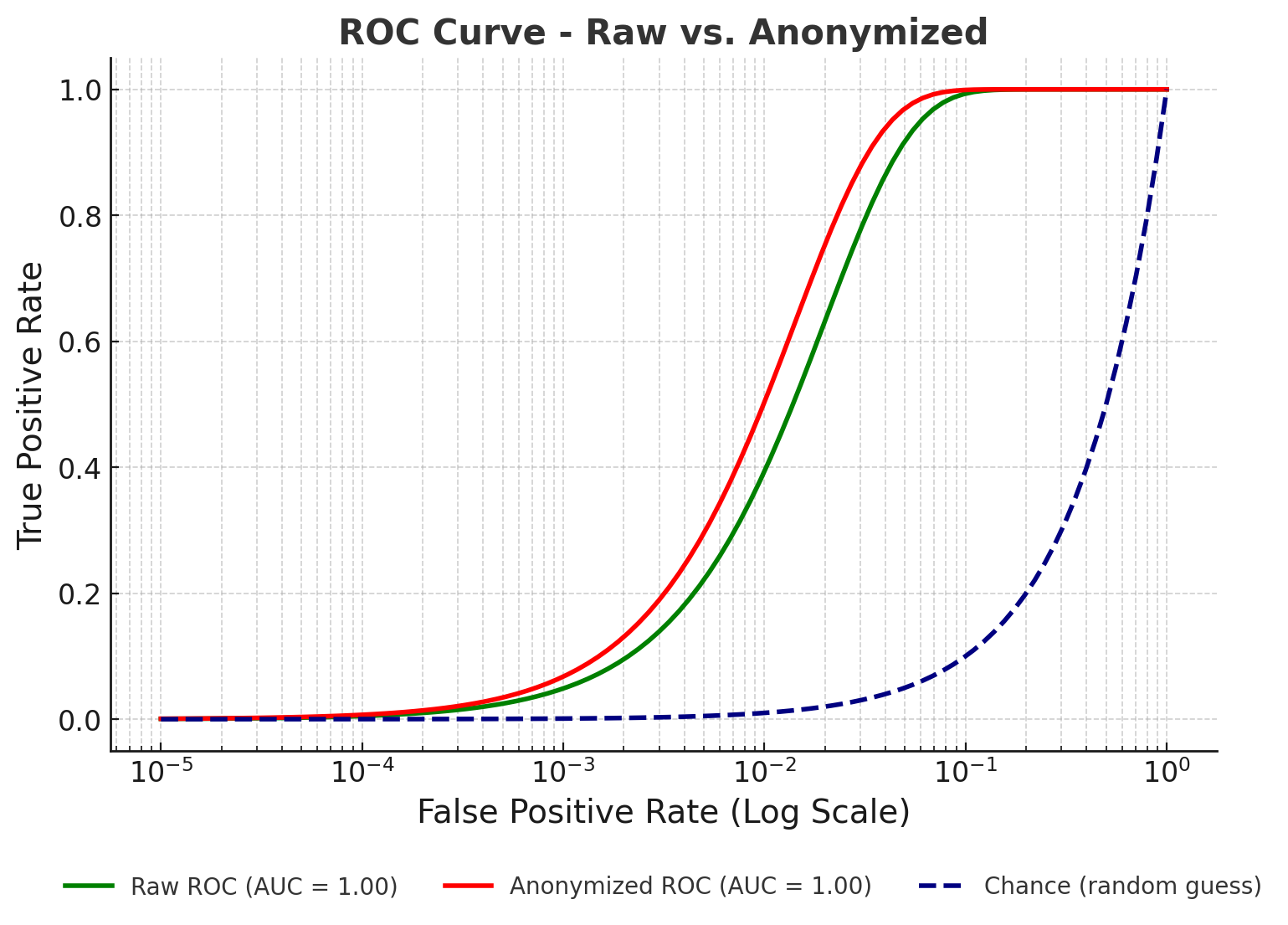}
\caption{\textbf{ROC curves comparing recognition performance between original faces and FaceAnonyMixer-protected faces}. The nearly identical AUC values confirm that verification accuracy is preserved after anonymization.}
\label{fig:ROC}
\end{figure}

\begin{figure}[htp]
\centering
\includegraphics[width=0.7\linewidth]{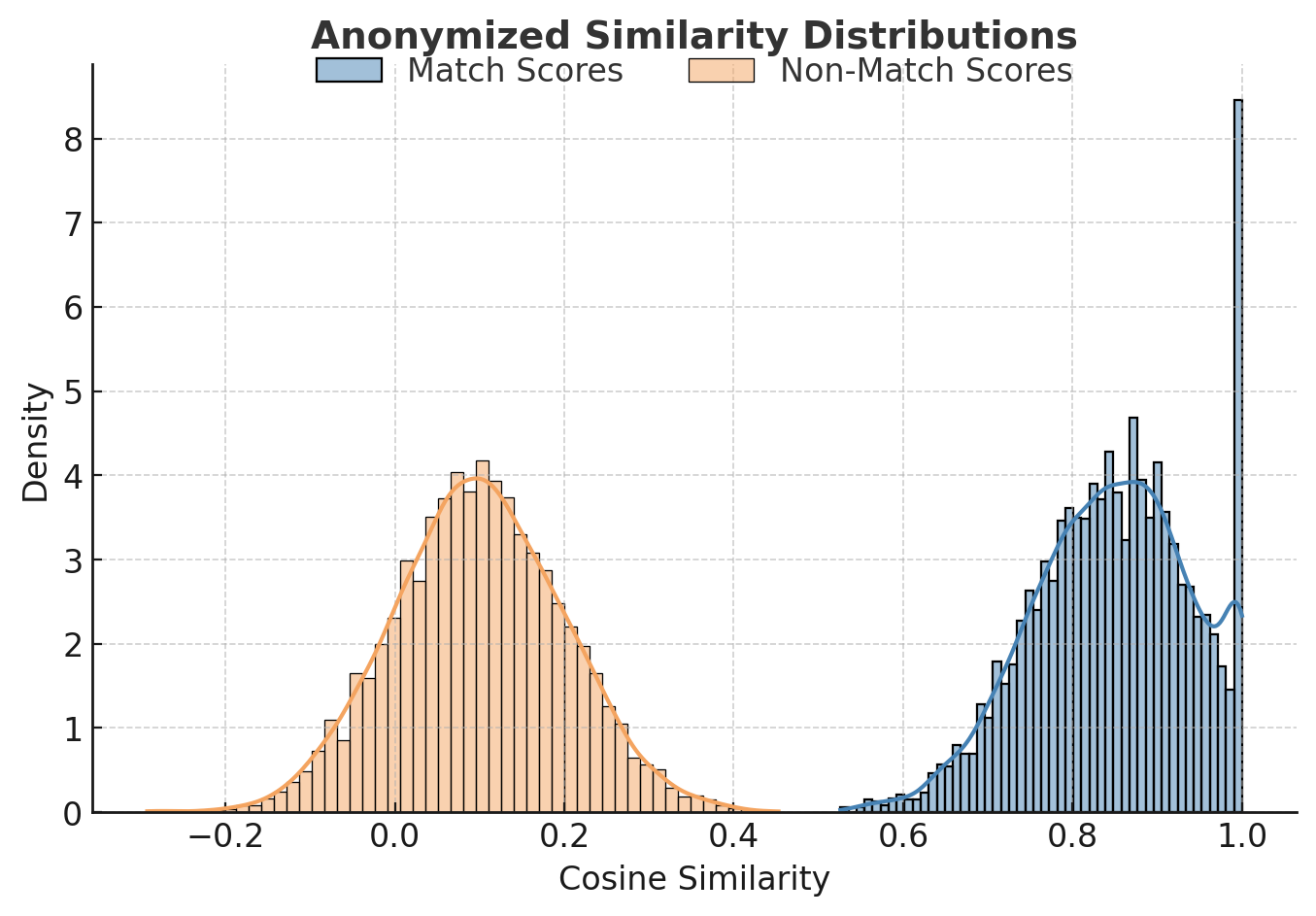}
\caption{\textbf{Cosine similarity distributions for match scores (same identity under the same key) and non-match scores (different identities under the same key)}. The absence of overlap indicates strong identity separability post-anonymization.}
\label{fig:sim}
\end{figure}

\begin{figure}[htp]
\centering
\includegraphics[width=0.7\linewidth]{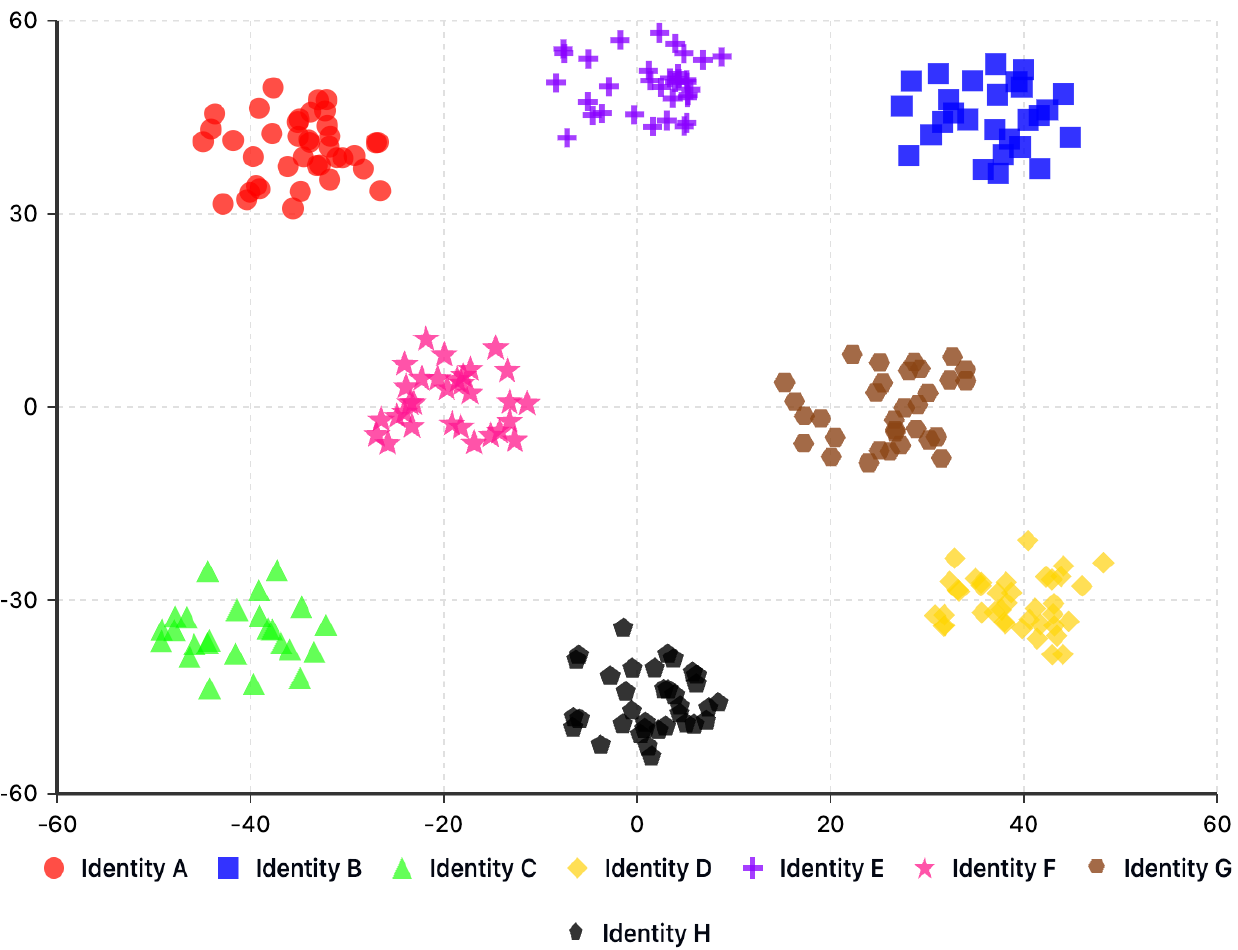}
\caption{\textbf{t-SNE visualization of ArcFace embeddings for FaceAnonyMixer-protected faces (single key)}. Clusters remain compact and well-separated by identity, demonstrating preservation of the recognition feature space.}
\label{fig:tsne_anon}
\end{figure}
\begin{figure*}[t]
\centering
\begin{minipage}[t]{0.32\textwidth}
    \centering
    \includegraphics[width=0.95\linewidth]{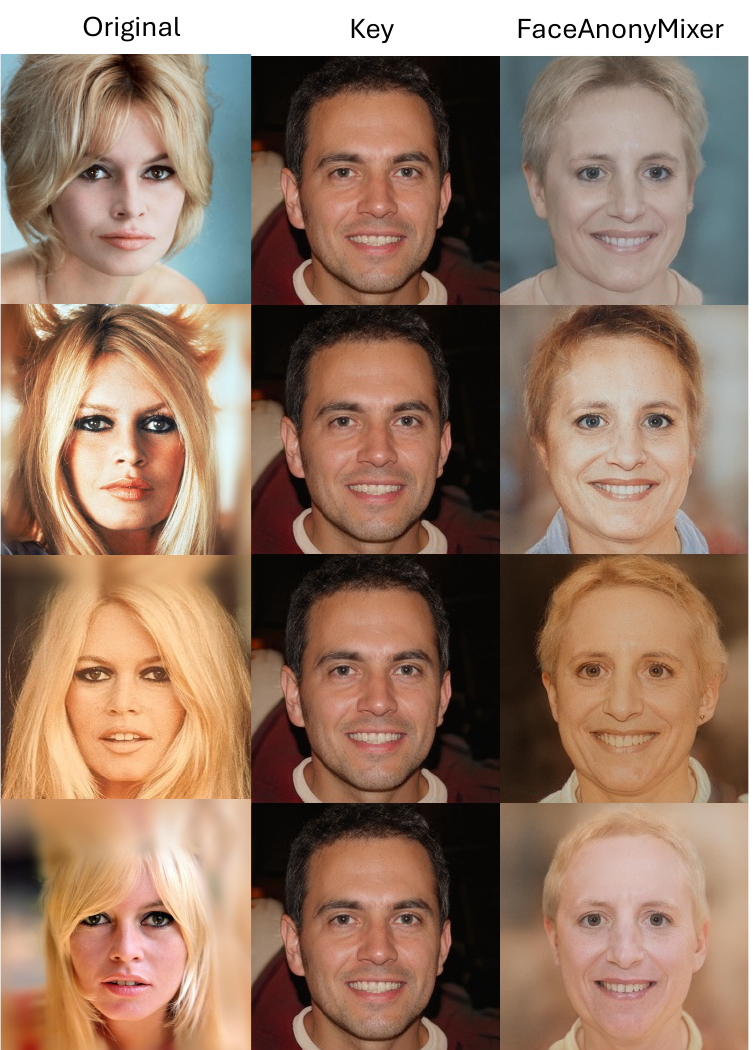}
    \caption{\textbf{Effect of removing the attribute-preserving loss.} Without attribute-preserving loss, anonymized faces lose key attributes such as expression and pose, reducing realism and utility.}
    \label{fig:attr}
\end{minipage}
\hfill
\begin{minipage}[t]{0.32\textwidth}
    \centering
    \includegraphics[width=0.95\linewidth]{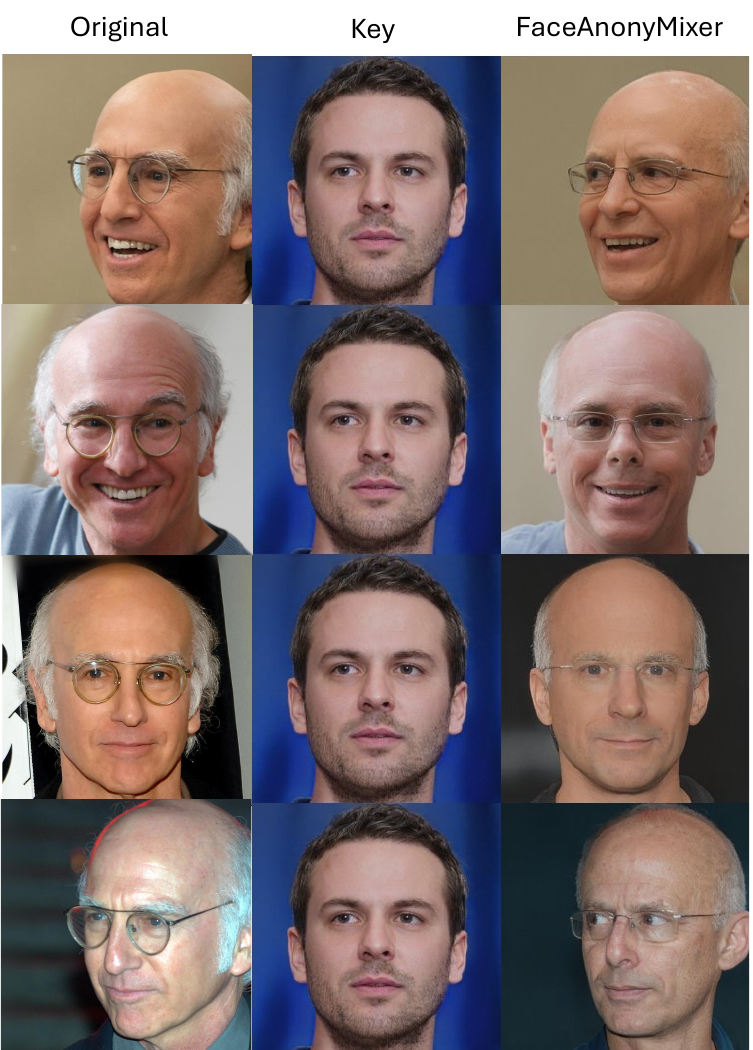}
    \caption{\textbf{Effect of removing the identity-preserving loss.} Without ID-preserving loss, anonymized images of the same individual under the same key become inconsistent, weakening verification performance.}
    \label{fig:cons}
\end{minipage}
\hfill
\begin{minipage}[t]{0.32\textwidth}
    \centering
    \includegraphics[width=0.95\linewidth]{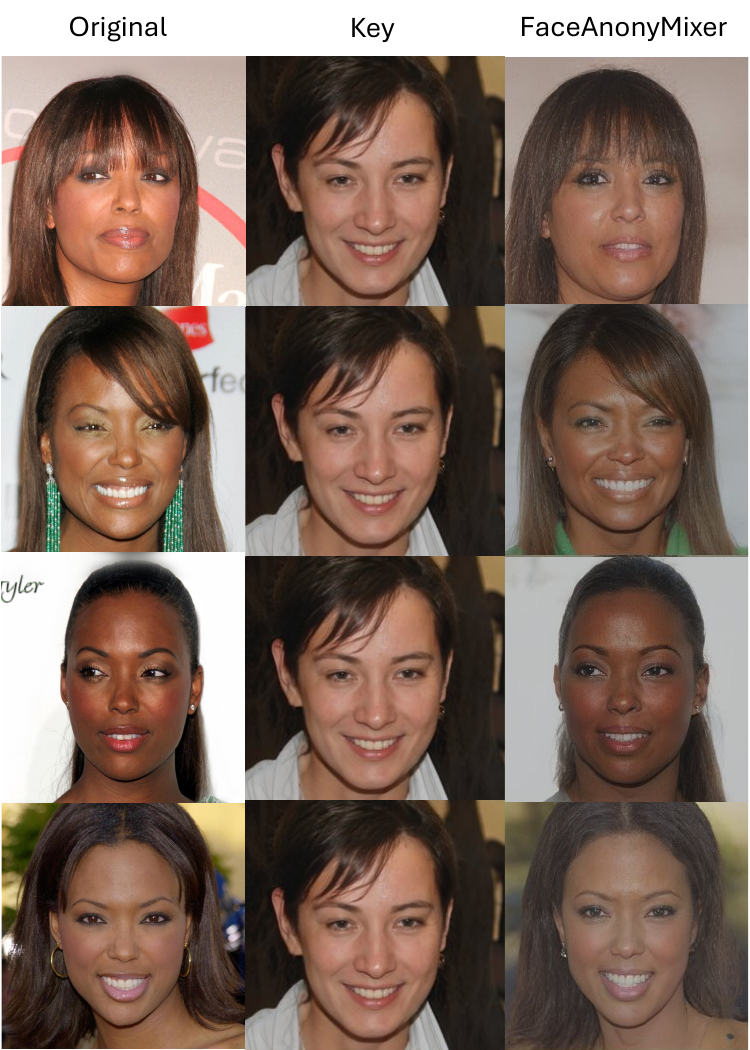}
    \caption{\textbf{Effect of removing the anonymity loss.} Residual identity features remain visible, compromising privacy and enabling partial re-identification.}
    \label{fig:anon}
\end{minipage}
\end{figure*}

\begin{figure}[htp]
\centering
\includegraphics[width=0.7\linewidth]{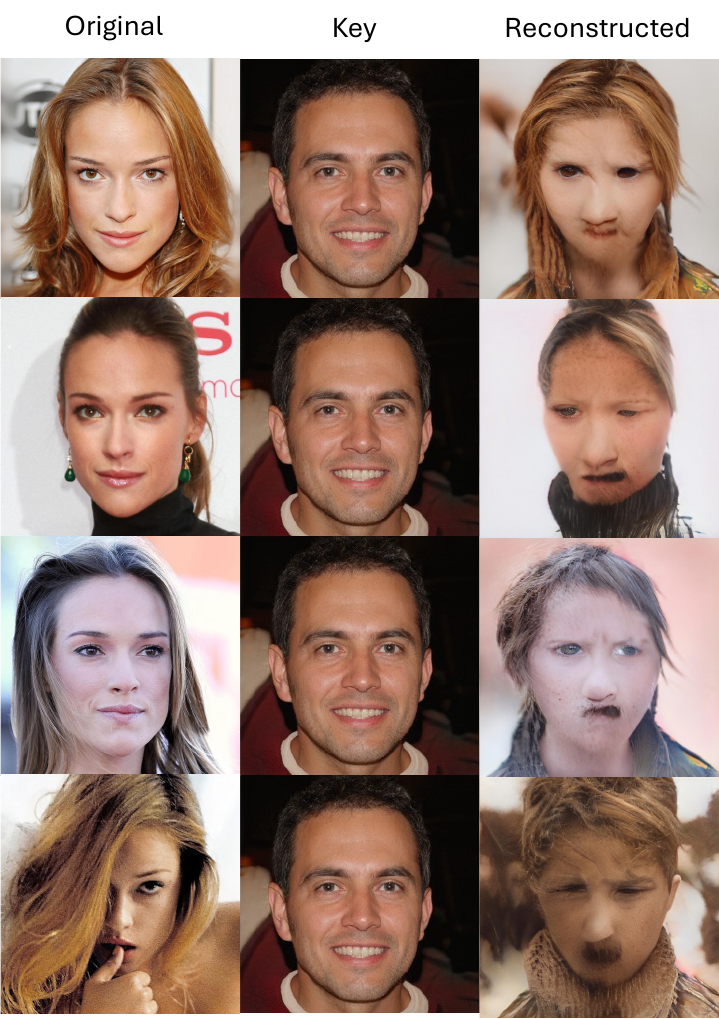}
\caption{Results of mapper-based inversion attacks on FaceAnonyMixer-protected faces. Reconstructed outputs fail to recover identity-specific traits, confirming strong irreversibility even under strong attack.}
\label{fig:recon}
\end{figure}

\begin{figure*}[t]
\centering
\includegraphics[width=0.595\linewidth]{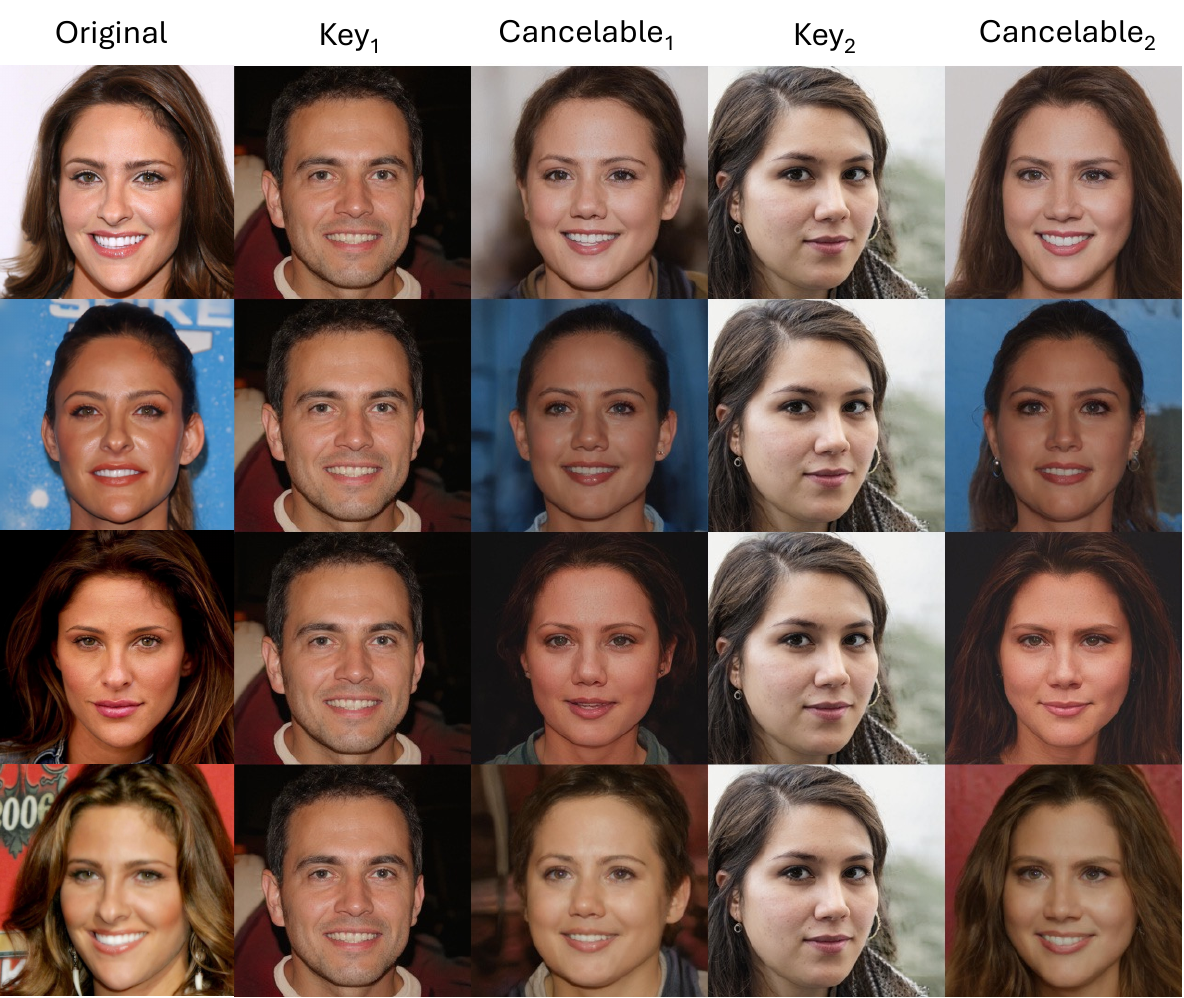}
\includegraphics[width=0.595\linewidth]{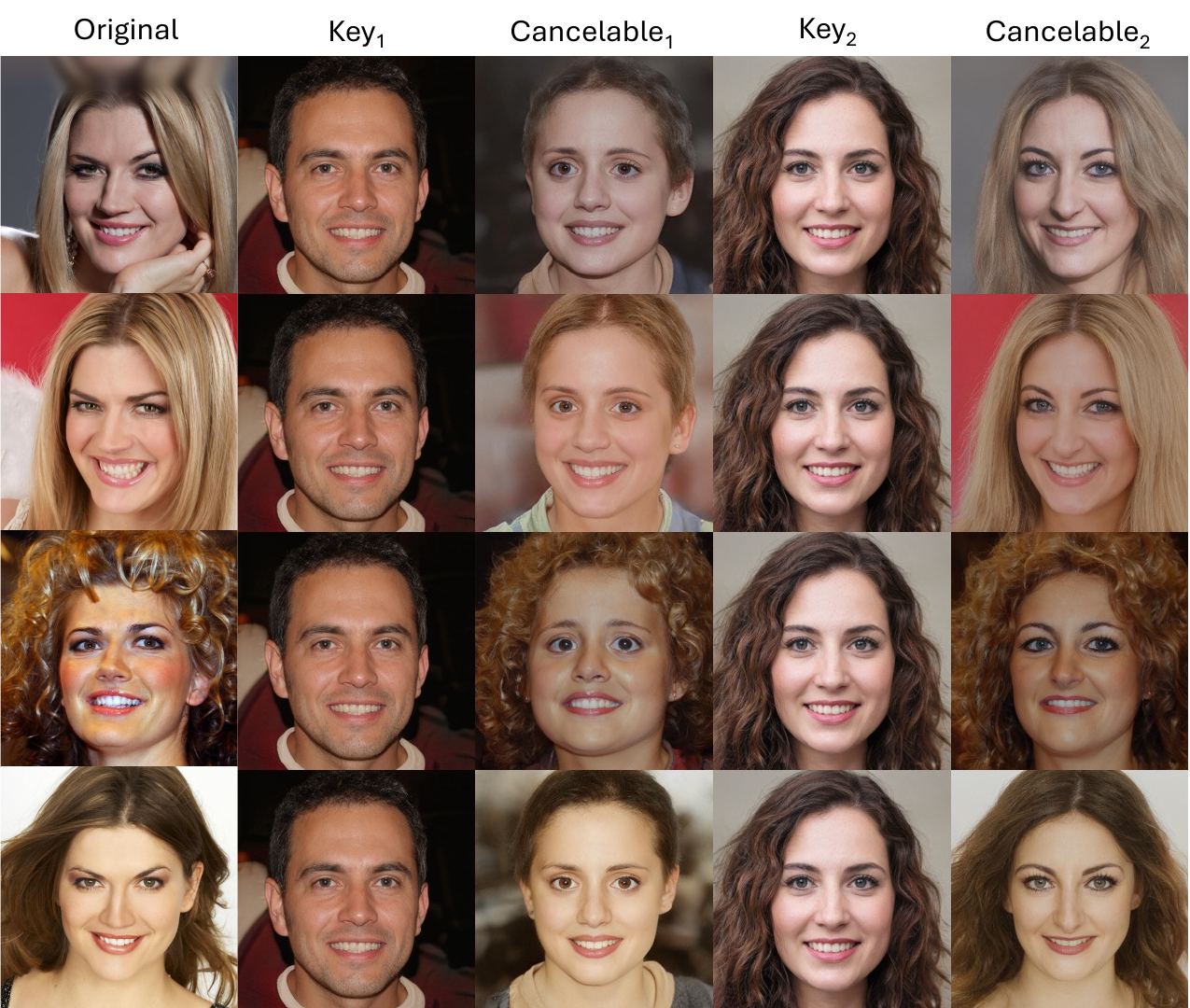}
\caption{Qualitative comparison of two subjects with two anonymization keys. FaceAnonyMixer outputs retain non-identity attributes (pose, expression, illumination) while achieving unlinkable transformations. This unified visualization demonstrates FaceAnonyMixer’s capability to maintain high‑fidelity attribute preservation alongside strong unlinkability across multiple anonymization keys.}
\label{multikey1}
\end{figure*}

\begin{figure*}[t]
\centering
\includegraphics[width=0.595\linewidth]{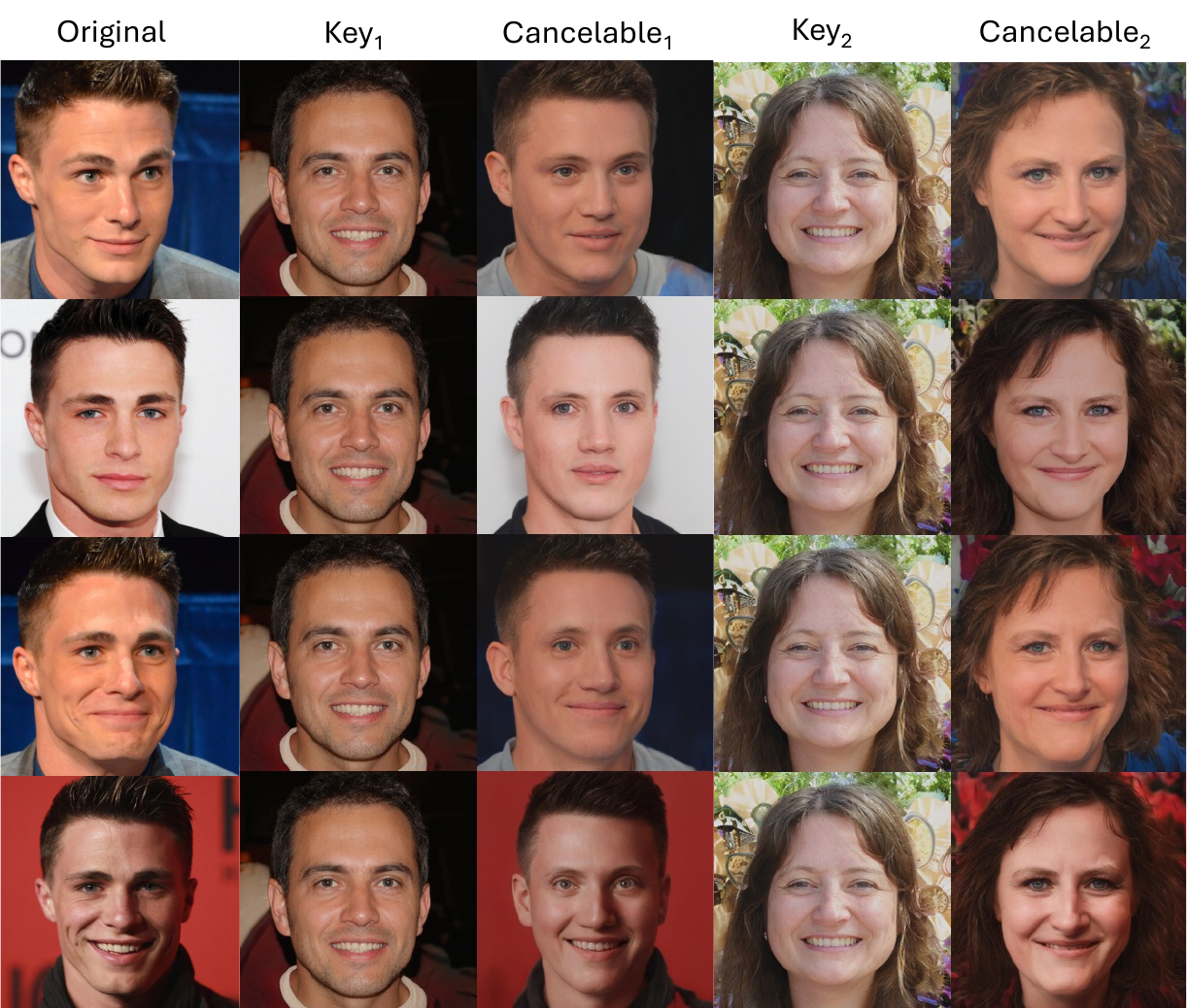}
\includegraphics[width=0.595\linewidth]{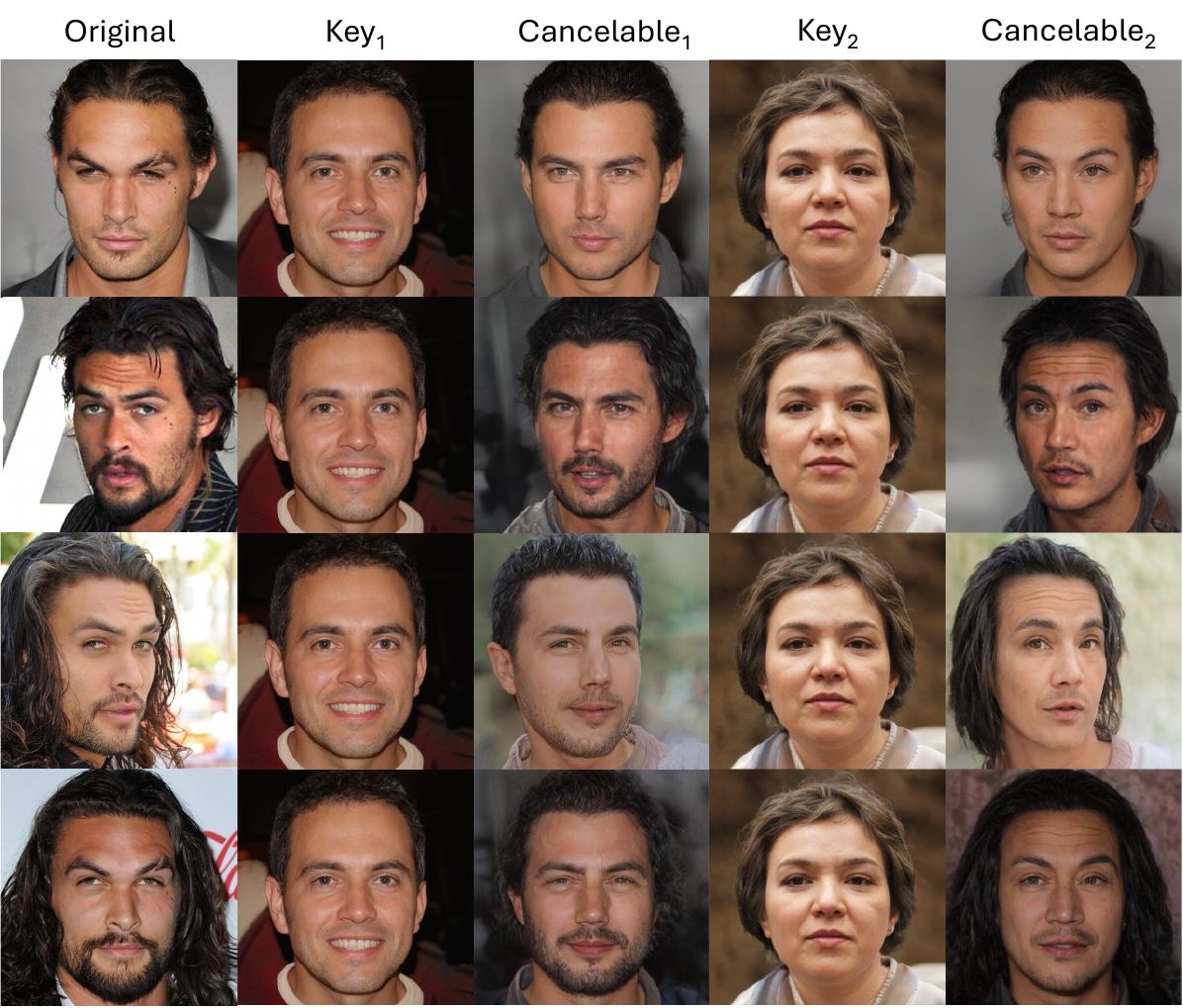}
\caption{Qualitative comparison of two subjects with two anonymization keys. FaceAnonyMixer outputs retain non-identity attributes (pose, expression, illumination) while achieving unlinkable transformations. This unified visualization demonstrates FaceAnonyMixer’s capability to maintain high‑fidelity attribute preservation alongside strong unlinkability across multiple anonymization keys.}
\label{multikey2}
\end{figure*}

\noindent \textbf{Recognition Consistency.}  
A fundamental property of any cancelable biometric scheme is maintaining recognition accuracy after anonymization. Figure~\ref{fig:ROC} presents the ROC curves for original versus FaceAnonyMixer-protected faces, showing nearly identical AUC values, thereby confirming that verification performance remains intact.
Figure~\ref{fig:sim} further illustrates cosine similarity distributions for (i) match scores, comparing two protected images of the same identity under the same key, and (ii) non-match scores, comparing protected images of different identities. The absence of overlap between these distributions confirms that FaceAnonyMixer preserves clear separation between same-identity and different-identity pairs, despite anonymization.

\noindent \textbf{Embedding Space Visualization.}  
To assess the structure of the recognition feature space, we visualize ArcFace embeddings of anonymized images using two-dimensional t-SNE. As shown in Figure~\ref{fig:tsne_anon}, FaceAnonyMixer-protected images cluster tightly by identity, while inter-class separation remains evident. This demonstrates that the relative geometry of the embedding space is preserved—images of the same person remain close together, whereas distinct identities remain well-separated, even after anonymization.

\noindent \textbf{Additional Ablative Results for Loss Components.} Figsures~\ref{fig:attr} to \ref{fig:anon} illustrate the critical role of each loss term in FaceAnonyMixer. Removing the attribute-preserving loss (Figure~\ref{fig:attr}) results in noticeable distortions in facial expressions, pose, and texture, reducing visual realism and usability. Excluding the identity-preserving loss (Figure~\ref{fig:cons}) leads to inconsistencies between anonymized images of the same individual under the same key, undermining verification reliability. Similarly, omitting the anonymity loss (Figure~\ref{fig:anon}) allows residual identity traits to persist, weakening privacy protection and enabling partial re-identification. Together, these results confirm that all three loss components are essential for maintaining a strong balance between privacy, identity consistency, and image quality.

\noindent \textbf{Robustness to Mapper-Based Inversion Attacks.} Figure~\ref{fig:recon} illustrates the results of inversion attempts where an adversary uses latent replacement or a learned image-to-image mapper to reconstruct the original face from its anonymized version. In all cases, the reconstructed images exhibit no meaningful resemblance to the source identity, confirming the irreversibility of our approach.

\iffalse

\begin{figure}[htp]
\centering
\includegraphics[width=0.7\linewidth]{figures/supp_attr.pdf}
\caption{\textbf{Effect of removing the attribute-preserving loss}. Without this term, anonymized faces lose key attributes such as expression and pose, reducing realism and utility.}
\label{fig:attr}
\end{figure}

\begin{figure}[htp]
\centering
\includegraphics[width=0.7\linewidth]{figures/supp_cons.pdf}
\caption{\textbf{Effect of removing the identity-preserving loss}. Anonymized images of the same individual under the same key become inconsistent, weakening verification performance.}
\label{fig:cons}
\end{figure}

\begin{figure}[htp]
\centering
\includegraphics[width=0.7\linewidth]{figures/supp_anonn.pdf}
\caption{\textbf{Effect of removing the anonymity loss}. Residual identity features remain visible, compromising privacy and enabling partial re-identification.}
\label{fig:anon}
\end{figure}
\fi

\noindent \textbf{Multi-Key Unlinkability and Attribute Preservation.}  Figure~\ref{multikey1} and Figure~\ref{multikey2} qualitatively illustrate FaceAnonyMixer’s ability to preserve non-identity attributes while achieving strong unlinkability across multiple keys. For each subject, the anonymized outputs generated using two different keys remain visually realistic, retaining critical attributes such as pose, facial expression, illumination, and occlusion. At the same time, identity-defining features are effectively altered, ensuring that protected templates from different keys cannot be linked. These visual results validate the method’s capability to balance high-fidelity attribute preservation with robust privacy protection.

\begin{figure*}[htp]
\centering
\includegraphics[width=0.7\linewidth]{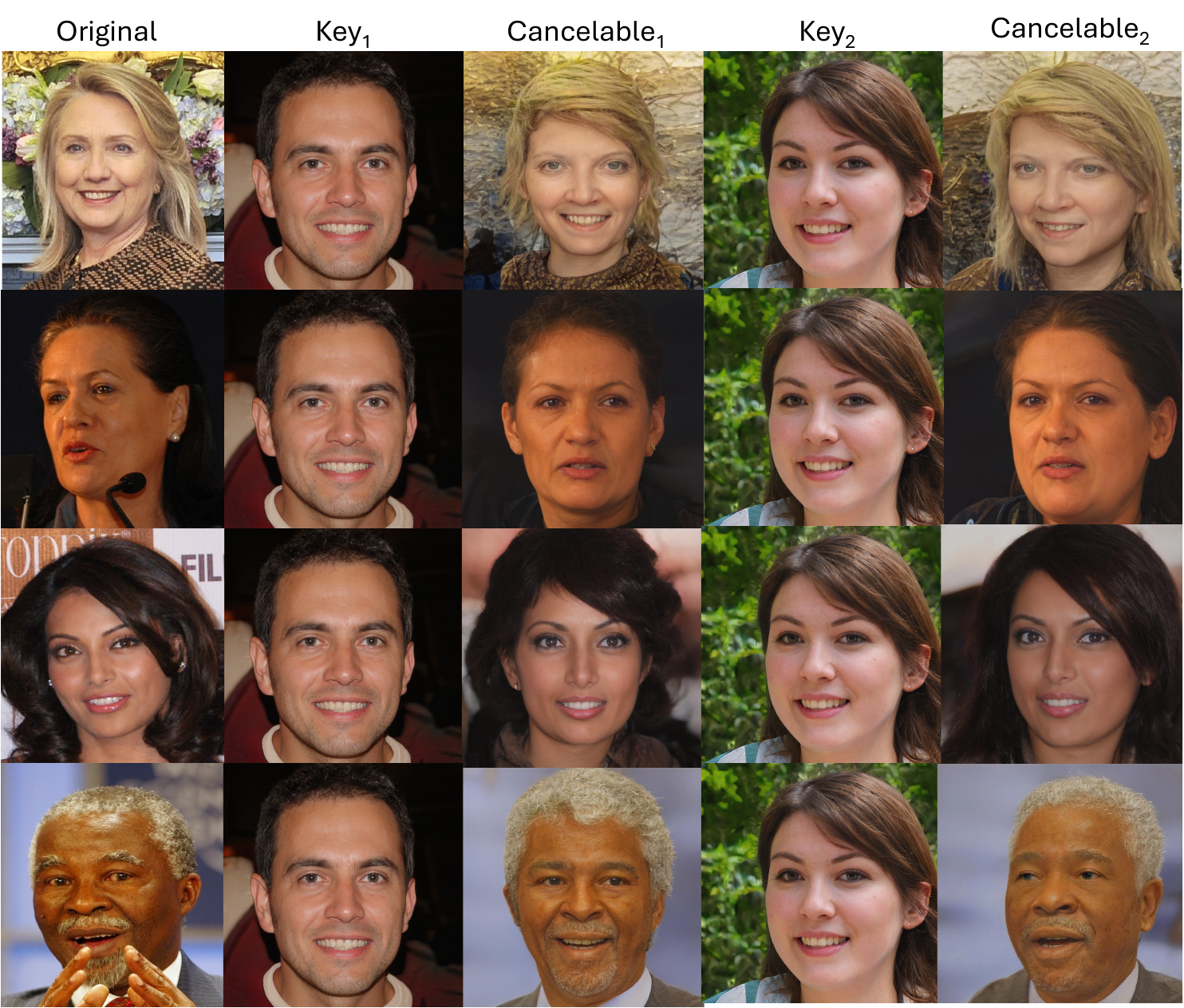}
\includegraphics[width=0.7\linewidth]{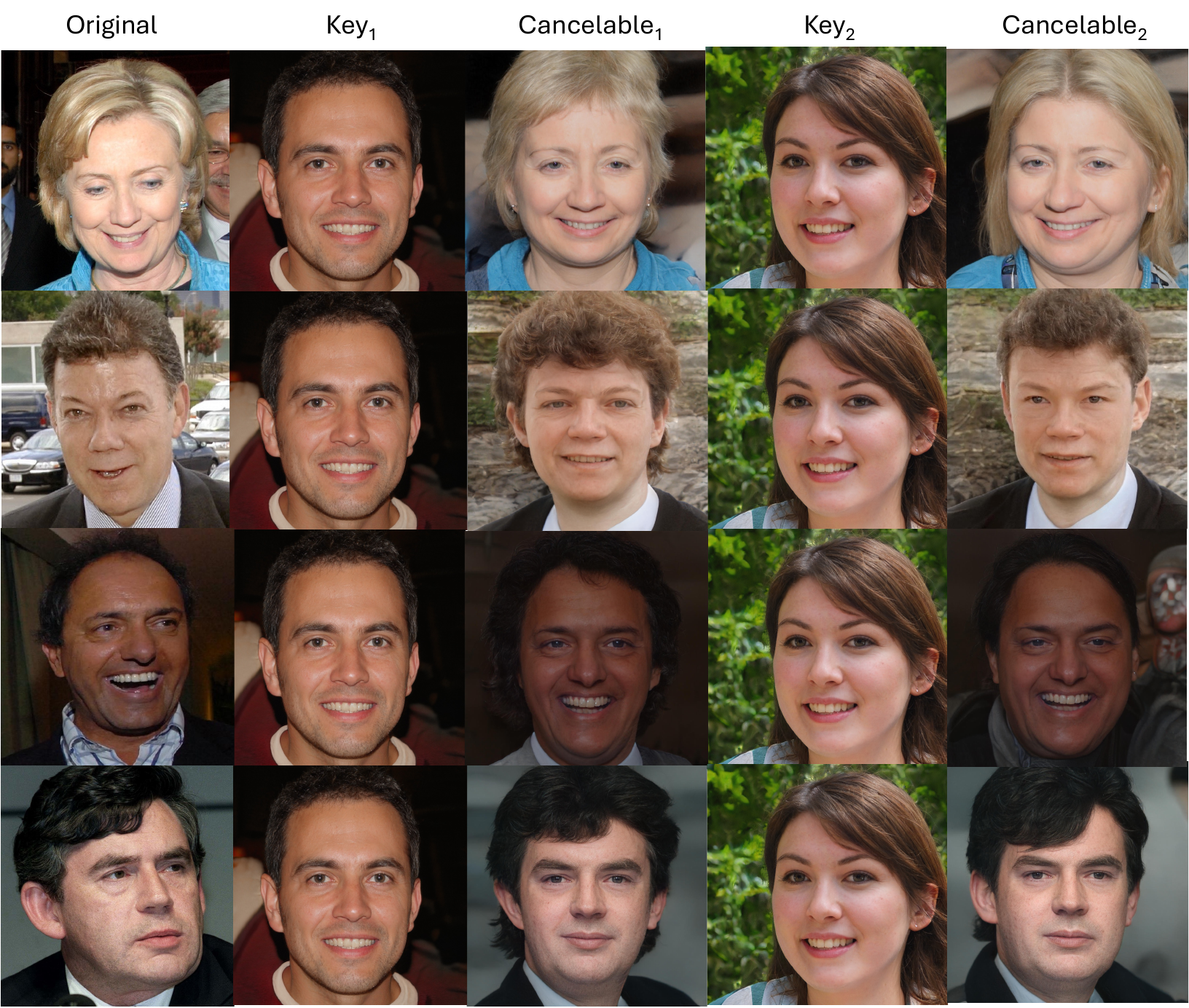}
\caption{Qualitative examples from the IJB-C dataset. FaceAnonyMixer preserves challenging attributes (pose, expression, and illumination) while anonymizing identity features, showing strong generalization to unconstrained conditions.}
\label{fig:ijbc}
\end{figure*}

\noindent \textbf{Qualitative Results on IJB-C Dataset [{\color{blue}23}].} 
Figure~\ref{fig:ijbc} presents qualitative examples of FaceAnonyMixer applied to the challenging IJB-C dataset, which includes images with significant variations in pose, illumination, expression, and occlusion. Our method generates visually realistic anonymized faces that faithfully preserve non-identity attributes such as pose and facial expression while effectively obfuscating identity-related features. The results demonstrate that FaceAnonyMixer generalizes well to unconstrained real-world conditions, maintaining both high visual fidelity and privacy protection across diverse scenarios.

%\end{document}

\end{document}